\documentclass{article}

\PassOptionsToPackage{numbers, compress}{natbib}

\usepackage[eandd, preprint]{WeGenBench}

\usepackage{hyperref}       
\usepackage{url}            
\usepackage{booktabs}       
\usepackage{amsfonts}       
\usepackage{nicefrac}       
\usepackage{microtype}      
\usepackage[table]{xcolor}  
\usepackage{amsmath}
\usepackage{multirow}
\usepackage{longtable}
\usepackage{graphicx}

\title{WeGenBench: A Multidimensional Diagnostic Benchmark towards Text-to-Image Model Optimization}

\author{
  Qian Liang$^{1, 3}$ \quad 
  Xiaomin Li$^{2, 3}$ \quad 
  Ying Zhang$^3$ \\
  Jia Xu$^2$ \quad
  Lihao Ni$^3$ \quad
  Hongrui Li$^3$ \quad
  Jingjing Li$^3$ \quad
  Jing Lyu$^3$ \quad
  Chen Li$^3$ \\
  \\ 
  $^1$University of Electronic Science and Technology of China \\
  $^2$Dalian University of Technology \quad
  $^3$Weixin, Tencent \\
  \texttt{202422080310@std.uestc.edu.cn} \quad
  \texttt{xmli22@mail.dlut.edu.cn} \quad
  \texttt{xjia@dlut.edu.cn} \\
  \texttt{\{yinggzhang, kumani, roryli, tinajjli, eckolv, chaselli\}@tencent.com}
}

\begin{document}

\maketitle

\begin{figure}[h]
  \centering
  \includegraphics[width=\textwidth]{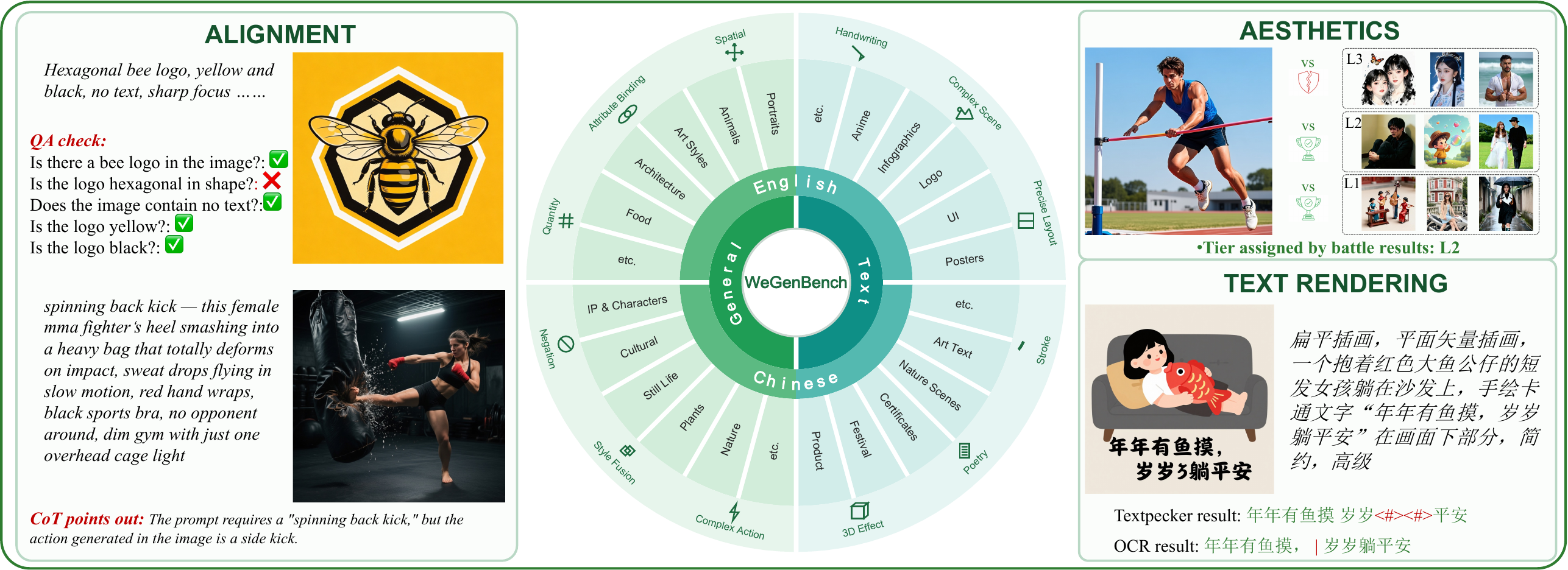}
\end{figure}

\begin{abstract}
  Recent text-to-image generation models have demonstrated remarkable capabilities in synthesizing highly realistic images from text inputs alone. Although existing benchmarks can evaluate the generation capabilities of various models to some extent, they struggle to comprehensively and accurately measure performance across multiple dimensions, often failing to reveal the inherent deficiencies of models in specific categories. To address these limitations, we propose WeGenBench, a novel benchmark designed for the comprehensive, multi-perspective evaluation of text-to-image generation capabilities. Our benchmark comprises a total of 4,000 test prompts across two primary categories, meticulously balanced between Chinese and English to evaluate bilingual and cross-cultural generation capabilities. Beyond macroscopic scene classification, we annotate each prompt with multi-dimensional tags tailored to the distinct content and challenges of each language, thereby refining the generation tasks into more specific sub-categories. Through a cross-dimensional evaluation mechanism leveraging both scene classifications and multi-dimensional tags, WeGenBench can precisely pinpoint model shortcomings in specific generation categories.  Furthermore, to measure generation quality more accurately, we design and validate several novel evaluation metrics by integrating Vision-Language Models (VLMs), which assess model performance on domain-specific tasks from three core aspects. Crucially, our approach yields both the assessment outcomes and the detailed reasoning trajectories, facilitating a rigorous verification of the accuracy and soundness of the evaluation results. Finally, we conduct systematic benchmarking on current state-of-the-art methods and provide an in-depth analysis of the limitations present in existing models.
\end{abstract}

\section{Introduction}

In recent years, text-to-image (T2I) generation models~\cite{podell2023sdxl,wu2025qwen,esser2024scaling,cai2025z,black-forest-labs-flux-2,cao2025hunyuanimage,nanobanana2,ernie-image,glm-image,gpt-image-2,diao2026sensenova,hidreamolimage} have demonstrated remarkable capabilities in synthesizing high-fidelity and controllable images. To reliably assess this rapid progress, numerous benchmarks and evaluation protocols~\cite{patel2024conceptbed,cho2023dall,hu2023tifa,zhu2023contrastive,hu2024ella,bakr2023hrs,tuo2023anytext,huang2023t2i,xu2023imagereward,li2026qwen,chang2025oneigbenchomnidimensionalnuancedevaluation} have emerged. However, existing benchmarks predominantly focus on either broad but coarse semantic coverage or highly specialized, single-scenario tasks. Moreover, they are almost exclusively English-centric, entirely overlooking the asymmetric challenges presented by different linguistic and cultural contexts---such as the stroke-level precision required for Chinese characters versus the complex typographic layouts demanded by long English texts. For instance, although benchmarks like PartiPrompts~\cite{yu2022scaling} explore diverse prompts, they are limited to scoring holistic capabilities and fall short in analyzing model performance on multi-dimensional tasks. Conversely, specialized benchmarks typically isolate specific generative dimensions; AnyText~\cite{tuo2023anytext} exclusively measures visual text rendering, while T2I-CompBench~\cite{huang2023t2i} isolates compositional alignment such as spatial relationships and attribute binding. While these specialized tools accurately measure performance on their respective narrow tasks, they inherently fail to capture the holistic capabilities of modern T2I models. Even combining multiple disparate benchmarks cannot yield a cohesive, multi-perspective diagnosis of a model's underlying strengths and vulnerabilities.

\begin{table}[h]
  \centering
  \small
  \renewcommand{\arraystretch}{1.2}
  \caption{Comparison of WeGenBench with existing text-to-image generation benchmarks. Our benchmark uniquely provides multi-dimensional multi-label tags and cross-dimensional scenario evaluations, comprehensively covering general semantics, aesthetic quality, and complex visual text rendering.}
  \label{tab:benchmark_comparison}
  \resizebox{0.95\linewidth}{!}{%
  \begin{tabular}{@{}lccccccc@{}}
    \toprule
    \textbf{Benchmark} & \textbf{General Semantics} & \textbf{Aesthetics Eval.} & \textbf{Text Rendering} & \textbf{Multi-dimensional Tags} & \textbf{Bilingual} & \textbf{VLM-based Metrics} & \textbf{Scale (Prompts)} \\
    \midrule
    PartiPrompts~\cite{yu2022scaling} & \checkmark & Partial & Partial & $\times$ & $\times$ & $\times$ & 1,600 \\
    TIFA~\cite{hu2023tifa} & \checkmark & $\times$ & $\times$ & Partial & $\times$ & \checkmark & 4,000 \\
    GenEval~\cite{ghosh2023geneval} & \checkmark & $\times$ & $\times$ & $\times$ & $\times$ & \checkmark & 3,200 \\
    HRS-Bench~\cite{bakr2023hrs} & \checkmark & $\times$ & $\times$ & $\times$ & $\times$ & \checkmark & $\sim$50,000 \\
    ConceptBed~\cite{patel2024conceptbed} & \checkmark & $\times$ & $\times$ & $\times$ & $\times$ & \checkmark & $\sim$2,800 \\
    T2I-CompBench~\cite{huang2023t2i} & \checkmark & $\times$ & $\times$ & $\times$ & $\times$ & \checkmark & 6,000 \\
    AnyText~\cite{tuo2023anytext} & $\times$ & $\times$ & \checkmark & $\times$ & \checkmark & $\times$ & $\sim$1,000 \\
    DALL-Eval~\cite{cho2023dall} & \checkmark & \checkmark & \checkmark & $\times$ & $\times$ & Partial & $\sim$500 \\
    Qwen-Image-Bench~\cite{li2026qwen} & \checkmark & \checkmark & Partial & \checkmark & \checkmark & \checkmark & 5,000 \\
    DPG~\cite{hu2024ella} & \checkmark & $\times$ & $\times$ & \checkmark & $\times$ & \checkmark & 1,000 \\
    OneIG-Bench~\cite{chang2026oneig} & \checkmark & \checkmark & Partial & \checkmark & \checkmark & \checkmark & 2,440 \\
    TIFF~\cite{wei2025tiif} & \checkmark & $\times$ & \checkmark & \checkmark & $\times$ & \checkmark & 5,000 \\
    PRISM-Bench~\cite{qian2025prism} & \checkmark & $\times$ & \checkmark & \checkmark & $\times$ & \checkmark & 700 \\
    CVTG-2K~\cite{du2025textcrafter} & $\times$ & $\times$ & \checkmark & $\times$ & $\times$ & $\times$ & 2,000 \\
    LongText-Bench~\cite{bai2024longbench} & $\times$ & $\times$ & \checkmark & $\times$ & \checkmark & $\times$ & $\sim$1,000 \\
    ChineseWord~\cite{wang2026chinese} & $\times$ & $\times$ & \checkmark & $\times$ & $\times$ & $\times$ & $\sim$1,000 \\
    \midrule
    \rowcolor{gray!10}
    \textbf{WeGenBench (Ours)} & \checkmark & \checkmark & \checkmark & \checkmark & \checkmark & \checkmark & 4,000 \\
    \bottomrule
  \end{tabular}%
  }
\end{table}

To bridge this critical gap, we propose WeGenBench, a comprehensive, bilingual, and scalable benchmark designed for the holistic evaluation of T2I generation capabilities (see Tab.~\ref{tab:benchmark_comparison} for a systematic comparison). Instead of merely expanding prompt quantity, we construct meticulously balanced test samples (comprising 2,000 Chinese and 2,000 English prompts) spanning diverse sub-categories to maximize real-world applicability across different user demographics. Crucially, we introduce a novel multi-dimensional tagging mechanism, where each of the 4,000 test prompts is assigned multi-dimensional labels characterizing its stylistic constraints, text rendering complexity, and potential generative bottlenecks. This dual-layered classification---combining macroscopic scenario categorization with microscopic capability tags---enables WeGenBench to deeply diagnose model flaws across specialized sub-domains. By doing so, our benchmark provides actionable, multi-dimensional insights to more comprehensively evaluate the generative capabilities and characteristics of models, facilitating targeted model optimization.

Additionally, existing evaluation metrics~\cite{ku2024viescore,hosseini2025t2i,ghosh2023geneval,chen2025multi,lu2023llmscore,lin2024evaluating,zhang2024learning,kirstain2023pick,wu2023human} often reflect image generation results from a single perspective, struggling to effectively assess model capabilities in more multi-dimensional scenarios. Although some methods provide a holistic scalar score to indicate overall generation quality, they function as ``black-box'' evaluators~\cite{hessel2021clipscore,xu2023imagereward} that lack detailed diagnostic interpretability. Consequently, they are largely blind to localized defects such as incorrect text spelling, structural deformations, or complex spatial misalignments, rendering them inadequate for multi-dimensional capability assessment. To accurately measure model performance on WeGenBench, we build upon relevant prior methods and integrate the latest Vision-Language Models (VLMs) to explore and validate multiple novel evaluation metrics. We design specific experiments to validate the effectiveness of our evaluation methods and provide an in-depth analysis of the strengths and weaknesses of different approaches. Ultimately, we decided to evaluate generation models from three core dimensions using relevant methods: 1) Semantic Alignment, which emphasizes the generation performance in common scenarios and the semantic consistency between images and prompts. We utilize both Question Answering (QA) check and Chain-of-Thought (COT) deduction to conduct complementary evaluations through additive and deductive scoring mechanisms, respectively, thereby effectively enhancing the accuracy of the large models; 2) Aesthetic Quality, which focuses on the visual appeal and overall quality of the generated images. For this aspect, we ultimately employ an anchor-based match grading method for evaluation, which effectively mitigates the hallucination phenomenon of Vision-Language Models while maintaining low computational costs; 3) Visual Text Rendering, which targets the model's ability to generate text within images, such as spelling correctness and the legibility of complete sentences. We employ both OCR models and VLMs for this assessment to multi-dimensionally determine the accuracy of the text within the images. Furthermore, our proposed evaluation framework offers enhanced interpretability by providing explicit rationales and multi-dimensional justifications for the assigned scores, thereby facilitating the rigorous verification of the evaluation accuracy.

Finally, we systematically compile the latest text-to-image generation models, including leading open-source models and state-of-the-art commercial models, to conduct comprehensive benchmarking on WeGenBench. By deeply analyzing the evaluation results across our multi-dimensional framework, we not only highlight the overarching strengths and weaknesses of these models but also reveal their prevailing shortcomings and inherent vulnerabilities in specific generative categories. This thorough diagnosis also provides a clear, actionable roadmap for future model optimization or targeted post-training strategies.

Our main contributions are as follows:
\begin{itemize}
    \item We introduce WeGenBench, a comprehensive, bilingual (Chinese and English), and multi-category text-to-image generation benchmark. It comprises 4,000 prompts designed to evaluate image generation capabilities from multiple perspectives, capturing the distinct linguistic and visual challenges of different cultures. Furthermore, our benchmark assigns multi-dimensional tags to every single test prompt, which serve to precisely pinpoint the specific capabilities or vulnerabilities of models in highly specialized scenarios.
    \item We propose a diverse suite of evaluation metrics tailored to accurately assess model generative performance across various domain-specific tasks. Beyond merely assigning scalar scores, our framework provides explicit rationales and multi-dimensional justifications, thereby ensuring high interpretability. Extensive experiments validate that these interpretable metrics exhibit a strong correlation with human perception.
    \item We conduct an extensive evaluation of numerous state-of-the-art methods using our benchmark, providing an in-depth analysis of their capabilities and limitations across all evaluated dimensions.
\end{itemize}

\section{Related Work}

\textbf{Text-to-image generation.} 
The evolution of text-to-image (T2I) generation has witnessed remarkable breakthroughs in recent years. Early pioneering works primarily relied on Generative Adversarial Networks (GANs)~\cite{goodfellow2014generative,karras2019style, kang2023scaling}. While demonstrating potential, they often suffered from mode collapse and struggled to scale effectively. Subsequently, autoregressive models~\cite{yu2022scaling} paved the way for large-scale, open-domain synthesis by treating image generation as a sequence modeling task. Recently, diffusion models~\cite{rombach2022high} have emerged as the dominant paradigm, achieving unprecedented levels of photorealism and semantic alignment. State-of-the-art diffusion models~\cite{podell2023sdxl,wu2025qwen,cao2025hunyuanimage} have demonstrated exceptional capabilities in generating high-fidelity images. Concurrently, several post-training methods and adapters~\cite{mou2024t2i,xue2025dancegrpo} have been proposed to enhance the generative capabilities of open-source models in specific scenarios. Nevertheless, as these models become increasingly sophisticated, accurately evaluating their nuanced capabilities within complex, domain-specific contexts---such as intricate spatial reasoning or dense text rendering---has emerged as a critical bottleneck in the current field.

\textbf{Benchmarks for text-to-image generation.}
As T2I models continue to advance rapidly, the development of comprehensive benchmarks has emerged as a pivotal research focus. Early evaluations predominantly relied on general-purpose image captioning datasets~\cite{lin2014microsoft}. Nonetheless, their simplistic, descriptive prompts fall significantly short of reflecting the sophisticated synthesis capabilities of modern generators. To address this, recent studies have introduced more targeted benchmarks. Some works concentrate on specific generative dimensions, such as spatial reasoning and attribute binding~\cite{huang2023t2i}, or exclusively focus on visual text rendering~\cite{tuo2023anytext}. Although effective within their specific domains, these narrowly focused benchmarks lack the breadth required to assess the comprehensive generation proficiency of modern models. Conversely, comprehensive evaluation frameworks explore diverse prompts but remain constrained by their limited scale or heavy reliance on costly human evaluation~\cite{yu2022scaling}. More recently, datasets incorporating human preference~\cite{xu2023imagereward,wu2023human} have been proposed, yet they focus heavily on general aesthetic appeal rather than multi-dimensional capability diagnosis. Crucially, existing benchmarks typically lack a hierarchical, multi-label tagging structure, making it difficult to pinpoint specific model vulnerabilities. Moreover, they predominantly focus on English prompts, largely ignoring the distinct linguistic barriers and cultural nuances inherent in diverse languages. For instance, the specific difficulties in rendering precise Chinese strokes fundamentally differ from managing intricate typographic layouts in lengthy English texts. This critical gap in bilingual, stress-tested, and multi-dimensional evaluation serves as the primary motivation behind WeGenBench.

\textbf{Evaluation metrics for text-to-image generation.} 
Closely tied to the construction of comprehensive benchmarks, designing robust and human-aligned quantitative metrics remains a core challenge. Early automated metrics primarily evaluate the statistical distance between generated and real image distributions~\cite{salimans2016improved,heusel2017gans}. By completely ignoring the conditioning text prompts, they are fundamentally incapable of measuring semantic consistency. To gauge text-image alignment, cross-modal retrieval models have been widely adopted~\cite{hessel2021clipscore}. However, these metrics often struggle significantly when processing complex attribute binding, spatial relationships, or localized defects~\cite{thrush2022winoground}. Recently, researchers have begun exploring the use of large Vision-Language Models as evaluators~\cite{ku2024viescore, lu2023llmscore, lin2024evaluating}. While these VLM-based metrics show immense potential in holistic comprehension, they frequently operate opaquely without providing interpretable, dimension-specific feedback. Additionally, their reliability is severely compromised by visual hallucinations, frequently leading them to falsely validate erroneous typography during complex text rendering assessments. To bridge this gap, we design a novel, multi-dimensional evaluation suite integrated within WeGenBench. By combining anchor-based aesthetic grading, a dual-track semantic alignment framework, and hallucination-resistant text metrics~\cite{zhu2026textpecker}, our approach effectively leverages advanced Vision-Language Models to provide highly interpretable, accurate, and localized diagnostic feedback across diverse generative domains.

\section{WeGenBench}

During the data construction process, WeGenBench evenly divides the test set into two core categories: General Image Scenarios and Text Rendering Scenarios. Within each category, the test cases are further equally split into Chinese and English prompts. Consequently, each category-language combination comprises 1,000 samples, resulting in a total of 4,000 samples for the entire benchmark. Meanwhile, to evaluate model capabilities in specific scenarios more meticulously, we subdivide each core category into several scenarios and assign multiple multi-dimensional tags to each test prompt. These scenarios and tags are designed to classify the visual elements or generation characteristics involved in the prompts. Through this cross-dimensional classification mechanism leveraging both scenarios and tags, we can evaluate and diagnose the models' generation capabilities more accurately and comprehensively.

\subsection{WeGenBench-General}

\begin{figure}[h]
  \centering
  \includegraphics[width=\linewidth]{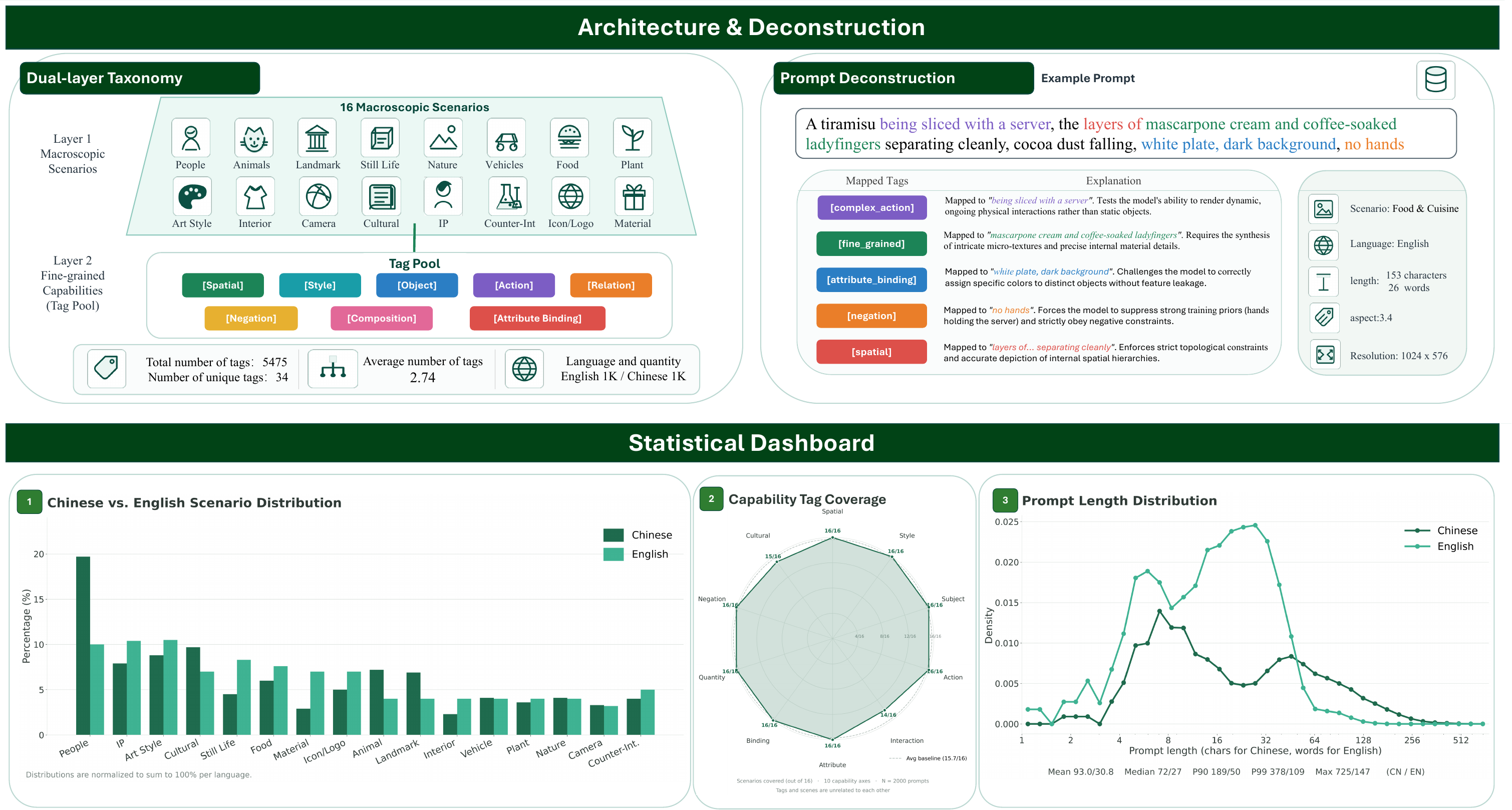} 
  \caption{\textbf{Overview of the WeGenBench-General benchmark}, detailing its architecture, deconstruction methodology, and selected statistics. \textbf{Top Left:} A dual-layer taxonomy comprising 16 macroscopic scenarios and a multi-dimensional capability tag pool. \textbf{Top Right:} The prompt deconstruction process, demonstrating how a complex prompt is mapped to specific capability tags (e.g., spatial, negation, and attribute binding) with detailed explanations of some tested model capabilities. \textbf{Bottom:} A statistical dashboard illustrating the dataset scale, language distribution, and capability tag coverage, specifically encompassing the distributional breakdown of scenarios and prompt lengths, as well as the cross-scenario coverage of representative tags across the 16 primary scenarios.}
  \label{fig:wegenbench_general}
\end{figure}

WeGenBench-General is a bilingual text-to-image evaluation benchmark comprising 2,000 meticulously curated prompts (1,000 for each language), and the relevant statistical information is shown in Fig.~\ref{fig:wegenbench_general}. Prompts in this category are primarily utilized for subsequent aesthetic quality and semantic consistency evaluations. To systematically assess models under real-world usage distributions, rather than relying on simple translations, the two subsets are independently sampled and annotated under a unified design objective. This approach authentically reflects the distinct expression habits of different user demographics---ranging from concise English instructions to complex, highly descriptive Chinese prompts. Ultimately, this dual-language, multi-dimensional design establishes a natural controlled setup, enabling a comprehensive evaluation of model robustness across diverse real-world generation tasks.

\textbf{Scenarios.} To ensure the diversity and comprehensiveness of the evaluation, we subdivide the general image category into 16 non-overlapping major scenarios based on real-world traffic distributions. These scenarios encompass Portraits \& People, Cultural Elements, Art Styles, IP \& Iconic Characters, Animals, Landmarks \& Architecture, Food \& Cuisine, Icons \& Logos, Still Life \& Commercial, Vehicles \& Machinery, Nature \& Weather, Counter-Intuitive scenes, Plants, Camera \& Composition, Materials \& Textures, and Interior \& Home. 
The distribution of these scenarios has been adjusted according to their varying usage frequencies.
For instance, the Chinese subset exhibits a higher concentration of Portraits (19.7\%) and Cultural Elements (9.7\%), aligning with the high-frequency demands of Chinese users. Conversely, the English subset presents a more balanced distribution, with no single scenario exceeding 11\%. Furthermore, both subsets deliberately retain samples with challenging tags such as Counter-Intuitive, Camera \& Composition, and Materials \& Textures to explicitly stress-test the models' spatial imagination, physical common sense, and texture reproduction capabilities. To ensure statistical stability, we enforce a strict minimum sample size for each scenario, guaranteeing reliable and variance-controlled metrics across all sub-domains.

\textbf{Style.} To evaluate the stylistic control capabilities of generation models, WeGenBench constructs a comprehensive and extensive style vocabulary. In the Chinese subset of the benchmark, there are 135 distinct styles. Notably, 63.5\% of the prompts do not specify a style, indicating a lack of strong stylistic constraints in these samples. Among the remaining 36.5\% of samples with explicit styles, the most frequent styles are: illustration (40), watercolor (18), 3D cartoon (15), oil painting (15), ink wash (14), cartoon (10), fine brushwork (9), comic (8), and Pixar style (8). Beyond these common categories, the vocabulary encompasses over 80 long-tail art types and genres. The distribution is broad, including traditional Chinese painters (e.g., Qi Baishi, Bada Shanren) and Western masters (e.g., Van Gogh, Monet, Picasso), as well as traditional craft styles like Dunhuang murals and paper cutting. Similarly, the English subset contains 143 styles, with 59.6\% of the prompts unspecified in style. The high-frequency styles in this subset include Art Deco, Watercolor, and Charcoal Drawing. Its long-tail distribution covers various Western art movements and subcultures, such as Bauhaus, Studio Ghibli, and Cyberpunk. This dual-layer coverage strategy of ``core high-frequency styles + long-tail art types and genres'' allows us to conduct fair horizontal comparisons across models on common styles, while also measuring the upper bounds of model generalization and transfer capabilities through rare long-tail styles.

\textbf{Prompt Length.} The average prompt length in the Chinese subset is 93.6 characters (median 73, range 2-736), with approximately 50\% of the samples falling between 51 and 200 characters. Short prompts ($\le$ 20 characters) and long prompts ($\ge$ 200 characters) account for 14.7\% and 9.5\%, respectively. For the English subset, the average length is 30.8 words (median 27, range 1-147); measured in characters, the average is 193. Approximately 47.1\% of the English samples fall between 21 and 40 words, and short prompts ($\le$ 10 words) account for 11.7\%.

\textbf{Capability Tags.} To support ``dimension-specific capability evaluation,'' we manually annotated each sample with a set of multi-label tags. The tagging systems for the Chinese and English subsets differ slightly to accommodate different linguistic habits: the Chinese version uses 21 open-ended phrases, aligning closer to practical operational tagging habits; the English version employs 13 controlled tags (e.g., \textit{fine\_grained}, \textit{attribute\_binding}), facilitating automated evaluation integration. Chinese samples carry an average of 2.78 tags, while English samples average 2.69 tags. Statistics indicate that both datasets prioritize ``fine-grained attributes,'' ``spatial relationships,'' and ``attribute binding'' as core evaluation dimensions, maintaining high consistency in core capability assessment. Beyond the core dimensions, the data distribution intuitively reflects the distinct expression preferences of the two user groups: the English version significantly emphasizes ``short prompts'' (22.7\%) and ``negation following'' (20.3\%), aligning with the tendency of English users to provide brief, objective statements; conversely, the Chinese version highlights ``explicit style descriptions'' (36.6\%) and ``complex interactions'' (18.5\%), matching the habit of Chinese users to employ extensive aesthetic modifiers to depict complex scenes. Overall, the differences in distribution proportions between the two datasets form a strong complementarity, effectively testing model robustness across two typical operational scenarios: ``short English inputs'' and ``complex, long Chinese descriptions.''

\textbf{Difficulty Levels.} To systematically quantify the complexity of the evaluation prompts and ensure a robust assessment of model capabilities, we categorize the WeGenBench-General prompts into five hierarchical difficulty levels (L1 to L5). This taxonomy evaluates the inherent complexity of each prompt based on two primary factors: the quantity of independent visualizable concepts and the density of specific capability constraints (e.g., spatial relationships, attribute binding, complex actions, and cultural elements). The levels are rigorously defined as follows: L1 (Minimal) involves 1-2 concepts devoid of modifiers or special constraints; L2 (Basic) encompasses 3-4 concepts with straightforward descriptions; L3 (Medium) includes 5-6 concepts coupled with 1-2 constraints; L4 (Hard) features 7 or more concepts integrated with multiple constraint combinations, characteristic of designer-level professional prompts; and L5 (Extreme) encompasses the most challenging scenarios, specifically those necessitating counter-intuitive reasoning, physical common sense, negation comprehension, complex multi-subject interactions, or intricate structural layouts. Based on our comprehensive evaluation across the 2,000 prompts, the overall difficulty distribution is as follows: L1 accounts for 6.3\%, L2 for 12.0\%, L3 for 24.4\%, L4 forms the largest segment at 39.7\%, and L5 constitutes 17.6\%. This distribution, significantly skewed towards the challenging L4 and L5 categories (combining for 57.3\%), ensures that WeGenBench-General serves as a rigorous stress test. It is deliberately designed to differentiate the nuanced capabilities of state-of-the-art models, transcending the evaluation of merely basic generation tasks.

\subsection{WeGenBench-Text}

\begin{figure}[h]
  \centering
  \includegraphics[width=\linewidth]{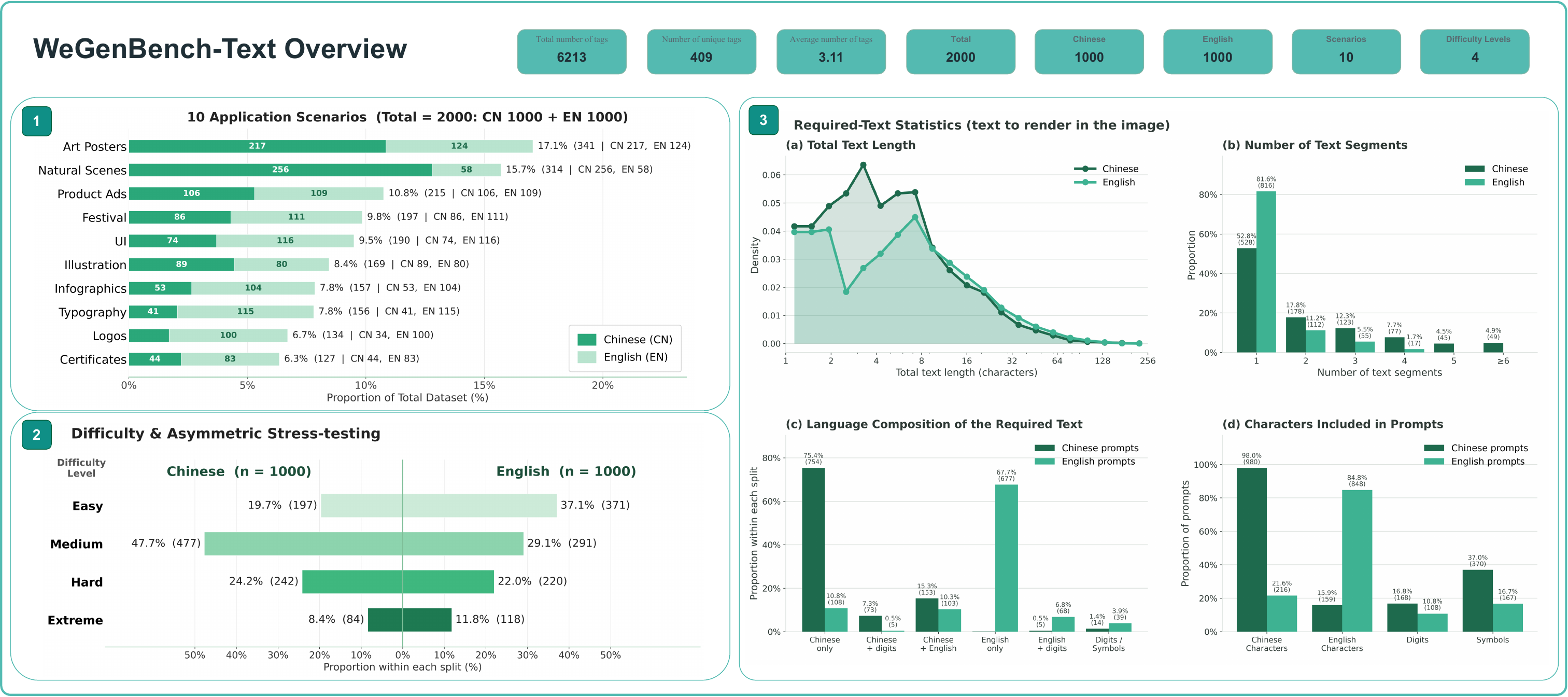} 
  \caption{\textbf{Overview of the WeGenBench-Text benchmark.} The dashboard details the basic information, tag distributions, and relevant statistics: (1) \textbf{Scenario Classification}: The prompts span 10 diverse scenarios, illustrating the distribution across different contexts. (2) \textbf{Difficulty Levels}: Displays the difficulty distribution of both Chinese and English prompts. (3) \textbf{Required-Texts Statistics}: Presents comprehensive statistics regarding the required text in rendering tasks, including variations in language, length, and style.}
  \label{fig:wegenbench_text}
\end{figure}

WeGenBench-Text is a bilingual visual text rendering benchmark comprising 2,000 meticulously curated prompts (1,000 for each language), with relevant statistical information illustrated in Fig.~\ref{fig:wegenbench_text}.

\textbf{Rendered Text Lengths.} Regarding the length of the required texts to be rendered, each Chinese prompt contains on average 21.5 characters, whereas each English prompt averages 22.4 characters. At the word level, each English prompt consists of 3.6 whitespace-delimited words on average, with the longest reaching 19 words. To systematically evaluate the text generation capabilities of large models, we bucketed the test samples based on the character length of their required text sequences. The Chinese subset presents a comprehensive and balanced distribution across length intervals: 330 short texts ($\le$ 6 characters), 269 medium texts (7--15 characters), 207 long texts (16--30 characters), and 194 ultra-long texts ($\ge$ 31 characters). Similarly, the English subset introduces a substantial proportion of challenging sequence lengths to stress-test the models, where long and ``ultra-long'' texts ($\ge$ 16 characters) combined amount to 498 instances (accounting for nearly 50.0\%), with the longest required text in a single prompt reaching 133 characters in total. The same heavy-tailed trend holds at the word level: 109 English prompts contain at least 9 words, of which 12 exceed 15 words. This carefully calibrated distribution directly targets the fundamental pain point of current Vision-Language Models, which are highly prone to spelling errors and omissions when rendering lengthy strings.

\textbf{Scenarios.} The Visual Text Rendering task remains a significant challenge in text-to-image generation, particularly in complex tasks involving non-English characters or dense multi-text distributions. We divide this portion of the test set into Chinese and English categories, each comprising ten major scenarios: natural scenes, artistic posters, product commercials, LOGOs and charts, UI interfaces, anime illustrations, popular science diagrams, certificates and medals, artistic typography, and festival scenes. These scenarios not only cover the vast majority of common situations encountered in daily life but also introduce fictional scenarios to evaluate the models' creativity and generalization capabilities. Based on these scenarios, WeGenBench can precisely assess a model's generation capability in specific scenarios, thereby providing clear directions for future model improvements.

\textbf{Text Types.} This dataset exhibits a rich distribution of linguistic components. The Chinese subset primarily focuses on Hanzi, authentically replicating common contexts in Chinese typography. Specifically, out of the 2,200 Chinese text segments, 1,754 segments (79.7\%) are purely Chinese, while mixed strings such as ``Chinese + digits'' (104 segments, 4.7\%) and ``Chinese + English'' (64 segments, 2.9\%) constitute most of the remainder. In contrast, the English subset introduces more complex character compositions. Among the 1,273 text segments, although 856 segments (67.2\%) are purely English, it still includes a large scale of cross-lingual and mixed strings, such as ``English + numbers'' (77 segments, 6.0\%) and ``Chinese + English'' (119 segments, 9.3\%). This is intended to test the robustness and cross-lingual transfer capabilities of generation models when rendering heterogeneous character sets. Moreover, stylized rendering and visual effects of text are another crucial dimension evaluated by this benchmark. Both subsets extensively cover generalization tests for basic fonts, such as ``handwriting'' (Chinese: 115, English: 79) and ``bold'' (Chinese: 110, English: 95). Meanwhile, a divergence exists in special visual effects: the Chinese subset places greater emphasis on three-dimensional/embossed effects (49 prompts, versus 23 in English), as well as on overall brushwork and design aesthetics, with 143 prompts explicitly requiring ``artistic typography.'' This setup enables multi-dimensional testing of the models' artistic style control capabilities across different linguistic character systems.

\textbf{Text Attributes.} Spatial layout control is a core challenge in text-to-image synthesis. This dataset evaluates models' spatial controllability through a ``precise positioning'' tag system. Statistics reveal a pronounced cross-lingual asymmetry in this dimension: 604 prompts (60.4\%) in the Chinese subset and 184 prompts (18.4\%) in the English subset explicitly specify the absolute physical location of the text in the image or the object medium it is attached to (e.g., ``located in the top right corner of the screen'' or ``printed in the center of the T-shirt''). This requires generation models not only to spell text correctly but also to possess precise cross-modal spatial anchoring and understanding capabilities. The ``precise typography'' metric directly measures a model's ability to handle complex topological structures in image space, such as alignment, intelligent line breaks, and multi-line spacing. Compared to the 224 (22.4\%) typographic constraints in the Chinese subset, the English subset's demand for multi-dimensional typographic control rises sharply due to the deliberate introduction of high-density text segments and extremely long string sequences. A total of 316 cases (31.6\%) contain strong instructions such as multi-line juxtaposition, center alignment, or list arrangements. This intensive typographic stress test poses severe challenges to existing models' structured composition and large-area text layout capabilities. To simulate visual noise and occlusions in the real world, we introduced a ``complex scenario'' control tag. The design proportions in this dimension are similar across both benchmarks: 265 cases (26.5\%) in Chinese and 240 cases (24.0\%) in English explicitly set cluttered background elements, foreground object occlusions, or strong perspective and lighting interference. Such complex settings significantly increase the difficulty of separating the generated text from the background and blending it naturally, serving as a direct test of aesthetic consistency and text usability in the image.

\textbf{Difficulty Levels.} Based on text length, the number of segments, and multiple spatial constraints, we systematically categorized the entire benchmark into four difficulty levels (Easy to Extreme). Due to the inherent complexity of writing Chinese characters, stroke errors easily occur even in short texts; thus, ``Medium'' (477 cases, 47.7\%) and ``Hard'' (242 cases, 24.2\%) constitute the main body of the Chinese test set, with ``Easy'' and ``Extreme'' accounting for 197 (19.7\%) and 84 (8.4\%) cases, respectively. The English subset exhibits a more dispersed distribution: ``Easy'' (371 cases, 37.1\%), ``Medium'' (291 cases, 29.1\%), ``Hard'' (220 cases, 22.0\%), and ``Extreme'' (118 cases, 11.8\%). Although a substantial share of English prompts are short, single-word strings that fall into the easier tiers, the deliberate introduction of extreme lengths and compound typographic constraints still lifts the English ``Extreme'' proportion (11.8\%) above that of Chinese (8.4\%), reflecting the ``asymmetric'' stress-testing design of our bilingual benchmark.

\section{Evaluation Metrics}

To comprehensively evaluate the generation capabilities of text-to-image models, we explored and analyzed numerous methods to propose a multi-dimensional evaluation strategy for WeGenBench. Our framework leverages the advanced reasoning capabilities of Vision-Language Models and is structured around three core dimensions: Alignments, Aesthetics, and Visual Text Rendering.

\subsection{Alignments}

\begin{figure}[h]
  \centering
  \includegraphics[width=0.95\linewidth]{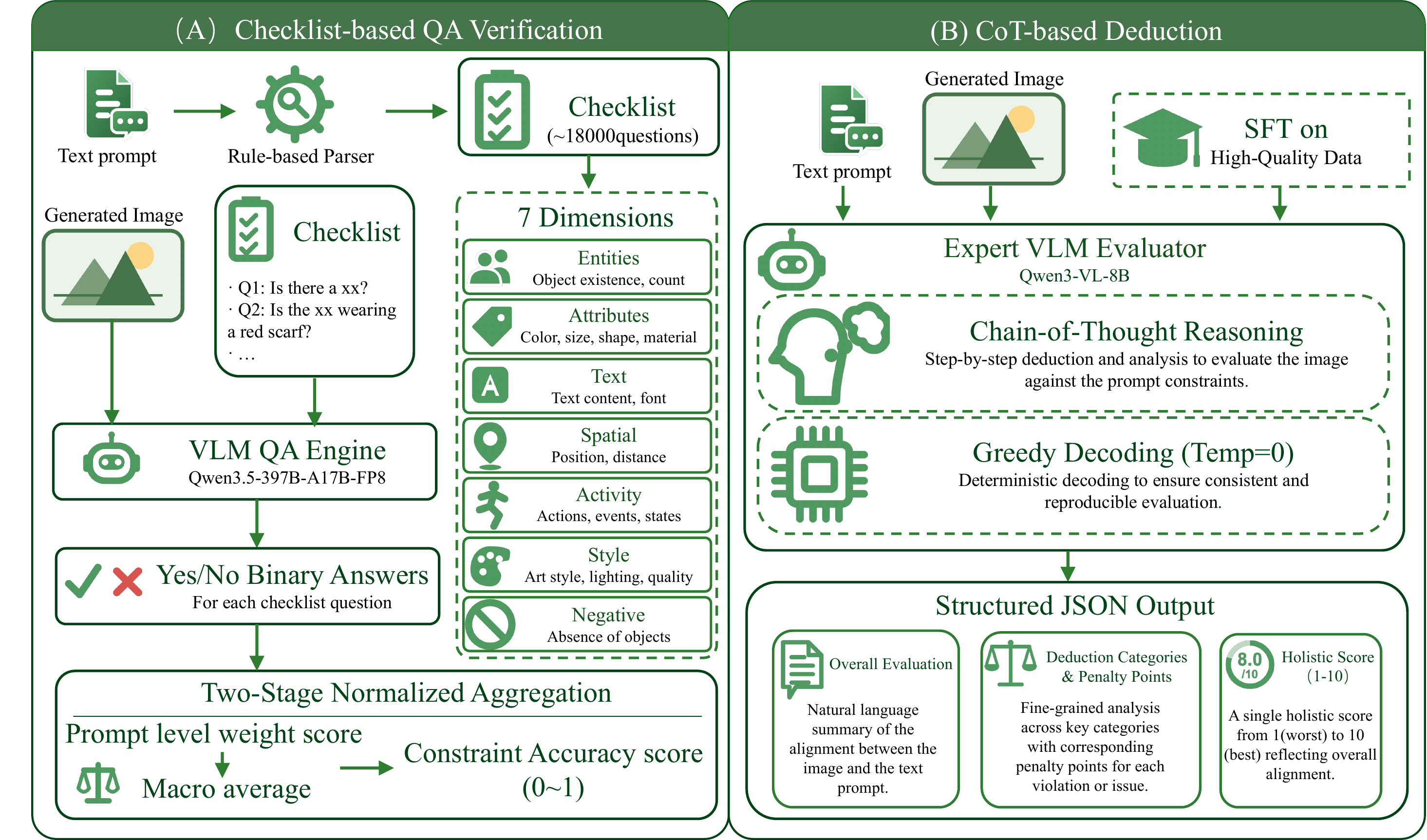} 
  \caption{\textbf{The overall pipeline of our Dual-Track Alignments Evaluation Framework.} The evaluation framework consists of two complementary branches. \textbf{(A) Checklist-based QA Verification} parses the text prompt into multi-dimensional checklists across 7 semantic dimensions. A VLM QA Engine then answers these binary questions based on the generated image, producing a Constraint Accuracy score through two-stage normalized aggregation. \textbf{(B) CoT-based Deduction} employs an Expert VLM Evaluator, fine-tuned on high-quality data, to perform step-by-step Chain-of-Thought reasoning. Using deterministic greedy decoding, it outputs a structured JSON containing an overall evaluation, specific deduction categories with penalty points, and a holistic score ranging from 1 to 10.}
  \label{fig:alignments_pipeline}
\end{figure}

Semantic alignment, defined as the degree to which a generated image faithfully reflects the conditioning text prompt, serves as a fundamental criterion for evaluating text-to-image generation models. Nonetheless, accurately measuring this alignment remains a complex challenge. Traditional cross-modal metrics like CLIPScore~\cite{hessel2021clipscore} often fail to capture multi-dimensional details, while directly prompting generic Vision-Language Models (VLMs) is prone to hallucinations and lacks diagnostic interpretability. Additionally, evaluation design faces an inherent trade-off: focusing too heavily on micro-level constraint verification can lead to a fragmented understanding of the image, whereas relying solely on macro-level semantic comprehension might overlook critical localized defects.

To address this dilemma, we propose an automated, VLM-based evaluation framework comprising two distinct, parallel, and complementary metrics (as illustrated in Fig.~\ref{fig:alignments_pipeline}): the Checklist-based QA Verification Metric and the COT-based Deduction Metric. Instead of fusing these approaches into a single opaque score, we utilize them as independent indicators. The QA metric acts as a rigorous inspector for micro-level details, while the COT metric serves as a holistic judge for global semantics and logic, jointly offering comprehensive insights into a model's alignment capabilities.

\subsubsection{Metric 1: Checklist-based QA Verification}

The first metric focuses on multi-dimensional, micro-level constraint verification. In this approach, the system decomposes the original prompt into a verifiable Yes/No Check Question List. Taking the triplet ``Prompt + Image + Question/Type/Weight'' as input, this metric guides the VLM to perform item-by-item visual question answering. Specifically, we construct a massive checklist comprising nearly 18,000 fine-grained questions. To systematically evaluate generation capabilities, we categorize over twenty distinct question types into seven macroscopic semantic dimensions. The distribution of these questions is meticulously designed: Entities (e.g., human, animal, object, plant, food, ip) account for $\sim$33\% of the questions; Attributes (e.g., color, counting, material, shape, size) account for $\sim$19\%; Text constraint questions comprise $\sim$11\%; Spatial/Relation (e.g., composition, location, relationship) account for $\sim$11\%; Visual Style (e.g., lighting, style, tone) make up $\sim$10\%; Scene (e.g., environment) represent $\sim$8\%; and Activity-related questions account for $\sim$6\%. Additionally, we include Negative constraints specifically tailored for English prompts. This hierarchical taxonomy ensures a comprehensive coverage of both fundamental objects and complex compositional relationships.

A critical challenge in this approach is the inherent variance in prompt complexity; the number of questions per prompt ranges from 2 to 20, and the sum of their independent weights varies significantly. To prevent complex prompts from disproportionately dominating the evaluation, we implement a two-stage normalized aggregation strategy. First, we calculate a normalized weighted score $\in [0,1]$ at the individual prompt level. Subsequently, we compute the macro-average across all prompts to derive the final Constraint Accuracy Score. This mechanism ensures statistical fairness while outputting detailed diagnostics, including dimension-specific scores and tag-level accuracy.

A key advantage of this metric lies in its high stability and strong interpretability. Once the checklist questions are fixed, the evaluation focal points remain constant across different models. By conducting itemized verification of explicit hard constraints (e.g., entity existence, specific colors, shapes, materials, quantities, and precise positions), the deduction rationale becomes highly explicit, facilitating the precise localization of specific generation defects. Furthermore, expanding evaluation dimensions or updating rules in the future is highly cost-effective. However, its primary limitation is that fragmenting the prompt into isolated, micro-level questions can easily lead to deviations in grasping the holistic semantics. Consequently, an image might score highly by satisfying all individual constraints while still appearing unnatural or logically flawed as a whole, occasionally resulting in a ``locally correct but globally suboptimal'' judgment.

\subsubsection{Metric 2: COT-based Deduction Review}

To evaluate the macroscopic semantic grasp that the first metric might miss and to provide a complementary perspective, the second metric employs a holistic review mechanism driven by Chain-of-Thought (COT) reasoning. In this approach, the system pre-constructs a unified Deduction Rule Prompt. Taking ``Prompt + Image + Deduction Rule'' as input, this metric requires the VLM to act as an impartial judge, conducting a comprehensive review from a global perspective based on established rules.

Instead of relying on generic VLMs acting as black-box judges, we specifically fine-tune a Vision-Language Model (e.g., Qwen-VL) using human-annotated ``expert review'' corpora. This Supervised Fine-Tuning (SFT) process enables the model to internalize human evaluation standards and stylistic nuances. Guided by a unified Deduction Rule Prompt and employing greedy decoding ($\text{temperature}=0$) to eliminate generative randomness, the model is strictly instructed to follow a structured reasoning paradigm. It outputs a highly interpretable text sequence: a comprehensive evaluation, followed by specific deduction categories with corresponding penalty points, and finally a holistic score ranging from 1 to 10.

To overcome the fluctuation issues typical of VLM evaluators, we introduce a rigorous Phase-1 Evaluator Validation mechanism. Before batch application, candidate checkpoints are benchmarked against human ground truth. We quantify their performance using a composite metric comprising $\pm1$ Accuracy, Category F1 score (for deduction dimensions), and Deduction Mean Absolute Error (MAE). Only the checkpoint that achieves optimal human alignment is selected for the final evaluation. This generative, CoT-driven approach exhibits a profound comprehension of global semantics, effectively compensating for the micro-level fragmentation of the QA Verification metric, while maintaining high interpretability, diagnostic granularity, and reproducible stability.

\subsection{Aesthetics}

\begin{table}[h]
  \centering
  \caption{A systematic comparison of the seven aesthetic evaluation schemes explored in our empirical study. We categorize them into two paradigms and analyze their mechanisms alongside fundamental trade-offs across interpretability (Interp.), stability, and critical bottlenecks. Our proposed \textbf{Level-wise Anchor-based Match} scheme provides the optimal balance.}
  \label{tab:aesthetic_eval_schemes}
  
  \resizebox{0.95\linewidth}{!}{%
  \renewcommand{\arraystretch}{1.2} 
  \begin{tabular}{@{}llp{3.8cm}ccp{4.2cm}@{}}
    \toprule
    \textbf{Evaluation Scheme} & \textbf{Paradigm} & \textbf{Core Mechanism} & \textbf{Interp.} & \textbf{Stability} & \textbf{Key Bottleneck (Cons)} \\
    \midrule
    Human ELO Match & Ranking & Human pairwise A/B testing & Low & High & First-glance bias, extremely high cost \\
    VLM ELO Match & Ranking & VLM reference-free match & Med. & Low & Forced fault-finding, misses minor defects \\
    Pair-wise Anchor Match & Ranking & VLM with predefined anchors (1v1) & High & Med. & Anchor selection cost, style mismatch bias, high VLM inference cost \\
    \rowcolor{gray!15}
    \textbf{Level-wise Anchor Match (Ours)} & \textbf{Ranking} & \textbf{VLM with adaptive anchors (1v1)} & \textbf{High} & \textbf{High} & \textbf{Anchor construction cost, bounded by VLM capability} \\
    \midrule
    VLM Rule-based & Scoring & Deductions via preset rules & High & Med. & Central tendency bias, sensitive to prompts \\
    VLM Open Obs. & Scoring & Describe-then-score & High & Low & Hallucinations, over-critical nitpicking \\
    VLM Multi-Choice & Scoring & Discretized sub-questions & V.~High & High & Information loss via forced discretization \\
    \bottomrule
  \end{tabular}%
  } 
\end{table}

Within the comprehensive evaluation framework for text-to-image generation models, the quantitative assessment of Aesthetic Quality has consistently been a highly challenging core issue. Aesthetic evaluation aims to measure the overall beauty of generated images across various scenarios---encompassing texture, composition, and color---which is highly constrained by the multidimensionality and subjective uncertainty of human perception.

To systematically address these evaluation challenges, we experimented with various aesthetic evaluation paradigms and conducted a comprehensive empirical analysis. These schemes can be broadly categorized into two primary paradigms: \textit{Relative Ranking-based Methods} and \textit{Absolute Scoring \& Aggregation-based Methods}. By analyzing their mechanisms, we uncovered fundamental trade-offs between human alignment, interpretability, and evaluation stability (summarized in Tab.~\ref{tab:aesthetic_eval_schemes}). Based on the insights gained from this exploration, we subsequently propose our \textbf{Anchor-based Match Grading} scheme, which serves as the primary aesthetic metric for WeGenBench.

\subsubsection{Empirical Analysis of Aesthetic Evaluation Paradigms}

The first paradigm, \textit{Relative Ranking-based Methods}, avoids absolute scoring by directly comparing images. We initially constructed a \textbf{Human ELO Match} evaluation baseline. This method conducts A/B testing through human pairwise comparisons and dynamically calculates the relative rankings of models using the ELO rating system (as defined in Eq.~\ref{eq:elo}). Following standard ELO rating practices, assuming the current ratings of models $A$ and $B$ are $R_A$ and $R_B$ respectively, the expected probability of model $A$ winning is defined as:
\begin{equation}
E_A = \frac{1}{1 + 10^{(R_B - R_A) / 400}}.
\label{eq:elo}
\end{equation}
After the match, model $A$'s rating is updated to $R_A' = R_A + K \cdot (S_A - E_A)$, where $S_A$ is the actual match result (1 for a win, 0 for a loss, and 0.5 for a draw), and $K$ is the update step size. While this scheme maximizes human advantages in aesthetic intuition and deformation detection, effectively avoiding the scale drift problem inherent in absolute scoring, it suffers from extremely high annotation costs. Moreover, human evaluators easily fall into the ``first-glance preference trap'' caused by strong stylization, and the method fails to provide multi-dimensional diagnostic explanations for reverse-engineering algorithm optimization. To reduce manual costs, we tested a \textbf{VLM ELO Match} scheme, utilizing Vision-Language Models to replace human scoring. Guided by prompts, Vision-Language Models can focus on specific dimensions and output judgment explanations, greatly improving evaluation efficiency. Nevertheless, our experiments revealed that current open-source Vision-Language Models still lack sufficient capability to detect minor deformations. When forced to determine a winner without a reference, logically inconsistent ``forced fault-finding'' or over-defensive phenomena frequently occur. To mitigate the reference-free issue, we further explored a \textbf{Pair-wise Anchor-based Match} scheme. This method transforms aesthetic evaluation into a structured feature comparison between the generated image and a meticulously curated library of anchor images by stitching them side-by-side. While it significantly improves interpretability and grounds the VLM's judgment, it remains highly susceptible to style mismatch bias when the generated image and the single anchor image differ drastically in content, and its stability is still compromised by the VLM's positional bias in 1v1 comparisons.

The second paradigm, \textit{Absolute Scoring \& Aggregation-based Methods}, focuses on decoupling perception and decision logic, directly assessing images based on predefined criteria. We first explored a lightweight \textbf{VLM Rule-based Scoring} scheme, translating experts' aesthetic criteria into structured penalty functions, formulated as Eq.~\ref{eq:rule_scoring}:
\begin{equation}
S(x) = S_{\mathrm{base}} - \sum_{i=1}^{k} \omega_i \cdot \mathrm{Penalty}_i(x),
\label{eq:rule_scoring}
\end{equation}
where $\mathrm{Penalty}_i(x)$ represents the deduction caused by the $i$-th detected flaw, and $\omega_i$ is a manually predefined weight reflecting the severity of that specific flaw. This method offers efficient inference and rapid iteration. Nonetheless, in practical testing, the score distribution is highly prone to central tendency bias (e.g., a massive concentration of images scoring 7--8). Moreover, the model is highly sensitive to prompts, and instruction fine-tuning easily leads to missed flaw detections or excessive deductions. Subsequently, we deeply validated two aggregation evaluation paradigms. The first is the \textbf{VLM Open Observation and Aggregation} scheme, adopting a ``describe-then-score'' strategy. The VLM first generates an open-ended feature set $\mathcal{F}$ containing the image's pros and cons, and subsequently aggregates this to derive the final score $S = g(\mathcal{F})$. This mechanism greatly enhances the logical consistency of model inference and can unearth deep details. However, experiments show that open observation easily induces the VLM into an over-critical ``nitpicking'' mode; visual hallucinations in long-text generation lead to a cascading accumulation of errors, and logical aggregation operators often struggle to capture the overall aesthetic essence of the image. To address this flaw, we tested the \textbf{VLM Multiple-Choice and Aggregation} scheme. This method decomposes vague aesthetic perception into objective sub-questions, forcing the VLM to make discrete decisions and aggregate scores via Eq.~\ref{eq:mc_scoring}:
\begin{equation}
S_{\mathrm{total}} = \sum_{j=1}^{M} \lambda_j \cdot \mathbb{I}(\mathrm{Choice}_j(x)).
\label{eq:mc_scoring}
\end{equation}
Here, $\mathbb{I}(\cdot)$ is an indicator function for the VLM's choice, and $\lambda_j$ represents the heuristically assigned score for each corresponding option. This approach effectively eliminates vague expressions and visual noise interference, possessing strong logical interpretability. Undeniably, forced discretized options inevitably cause a loss of perceptual information, often leading to extreme final scores, and a fixed option pool struggles to comprehensively cover all unknown generation defects.

Our empirical analysis indicates that no single conventional metric serves as a perfect solution. Ranking methods offer relative reliability but struggle with multi-dimensional defect attribution, whereas scoring methods provide detailed interpretability but are highly susceptible to VLM cognitive biases.

\subsubsection{Our Approach: Adaptive Level-wise Anchor-based Match Grading}

\begin{figure}[h]
  \centering
  \includegraphics[width=0.95\linewidth]{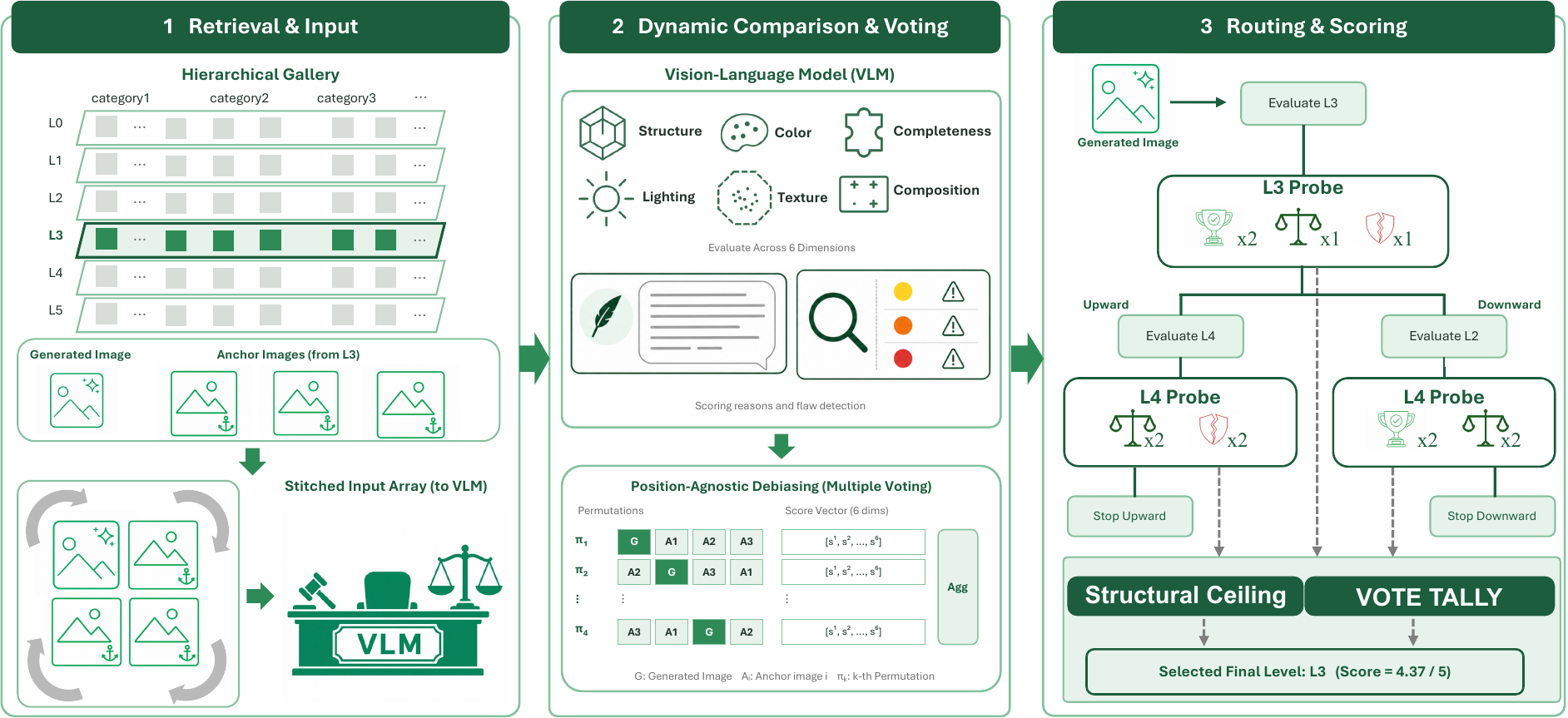} 
  \caption{\textbf{Overview of the proposed Adaptive Level-wise Anchor-based Match Grading pipeline for aesthetic quality evaluation.} The framework operates through three sequential stages: (1) \textbf{Retrieval Stage}: Based on the input image and text, semantic retrieval is performed against a Hierarchical Gallery (L0-L5) constructed via semantic categorization and representative curation. (2) \textbf{Dynamic Comparison Stage}: The input image is compared with selected anchor images. The system dynamically schedules the evaluation trajectory: routing upward (L4/L5) if the quality is evaluated as ``Above'' the anchor, routing downward (L0/L1/L2) if ``Below'', or terminating if it ``Meets'' the standard. (3) \textbf{Scoring Stage}: The final aesthetic level is determined by aggregating the comparison results through confidence weighting and average scores, while strictly applying a Structural Ceiling to ensure physical and structural correctness.}
  \label{fig:score_aes}
\end{figure}

To overcome the inherent limitations of conventional evaluation paradigms---specifically, to mitigate subjectivity while maintaining high interpretability and stability---we propose an \textbf{Adaptive Level-wise Anchor-based Match Grading} framework (illustrated in Fig.~\ref{fig:score_aes}). Instead of relying on reference-free absolute scoring or simple 1v1 comparisons, our method transforms aesthetic evaluation into a structured, multi-dimensional feature comparison between the generated image and a meticulously curated library of anchor images.

\textbf{Overview of the Evaluation Pipeline.} The evaluation operates as a dynamic, multi-stage tournament structured into three distinct phases. First, in the \textit{Retrieval Stage}, upon receiving an input image, the system retrieves a set of candidate anchor images from a hierarchical library (categorized from Level 0 to Level 5 across ten overarching semantic domains) based on semantic matching. Next, in the \textit{Dynamic Comparison Stage}, the input and a selected anchor image are stitched into a single side-by-side composite and fed into the Vision Large Language Model (VLM). Acting as an impartial judge, the VLM outputs a three-class relative judgment (\textit{below}, \textit{meets}, or \textit{above}) across six fine-grained dimensions (structure, texture, lighting, color, composition, and completeness) as well as an overall decision. Based on these ongoing comparison results, the system dynamically updates and selects subsequent anchor images to refine the comparison trajectory and determine the final tier for the generated image. Finally, in the \textit{Scoring Stage}, an aggregation module synthesizes all match outcomes. It computes the final aesthetic grade by comprehensively weighing the matched anchor tiers and applying a strict structural ceiling based on the image's structural integrity. To ensure the robustness, fairness, and accuracy of this pipeline, we implement three critical mechanisms and a rigorous gallery construction process:

\textbf{Hierarchical Anchor Gallery Construction.} To establish an authoritative and unbiased evaluation baseline, we construct a multi-dimensional anchor gallery. First, to eliminate cross-domain comparison biases, we map all test samples into ten macroscopic semantic domains. Within each domain, we define six quality tiers ranging from Level 0 (severe structural failure) to Level 5 (top-tier aesthetic quality). For each specific intersection of domain and quality level, human experts meticulously selected exactly three representative images as anchors. These 180 manually curated anchor images strictly define the structural and aesthetic standards for each level, serving as the absolute ground truth for subsequent VLM judgments.

\textbf{Adaptive Routing for Efficiency.} Exhaustively comparing an input against all anchor levels is computationally prohibitive. Our adaptive routing mechanism solves this by initiating a probe at the intermediate Level 3 (L3). Rather than relying on a single deterministic outcome, the trajectory is dynamically scheduled based on the distribution of multiple position-variant votes: to ensure conservative quality control, the downward threshold is highly sensitive, triggering a downward probe (progressively L2, L1, and L0) if there is at least one \textit{below} vote. Conversely, the upward probe (L4 and L5) requires stronger consensus, triggering only if there are at least two \textit{above} votes. Notably, in cases of severe positional conflict where both \textit{below} and \textit{above} votes coexist, the system simultaneously explores both downward and upward paths to collect comprehensive evidence. This dynamic scheduling ensures that computational resources are concentrated on the most informative boundary matches while preventing premature termination on ambiguous samples.

\textbf{Multi-Dimensional Stitched Comparison and Debiasing.} Building upon the curated hierarchical gallery, our Level-wise approach abandons isolated evaluation. Instead, it stitches the generated image alongside a category-matched anchor image (a 1v1 configuration) into a side-by-side frame. This forces the VLM to perform a direct, localized visual comparison against a domain-specific reference representing a unified quality baseline. Additionally, we explicitly decouple structural correctness from aesthetic appeal. The \textit{structure} dimension (evaluating anatomical deformations, text gibberish, or physical impossibilities) is granted veto power. To neutralize the well-documented positional bias in Vision-Language Models, we introduce a rigorous \textit{Position-Agnostic Debiasing} mechanism. For each level probe, the system executes multiple position variants (dynamically swapping the left and right positions). The valid votes are then aggregated: if a clear majority exists, it determines the decision; if there is a tie between adjacent decisions (e.g., \textit{below} and \textit{meets}), the system conservatively adopts the lower one; for non-adjacent divergences (e.g., \textit{below} and \textit{above}), it computes an average score to reach a balanced conclusion.

\textbf{Conservative Aggregation via Structural Ceiling.} The final quality level is not a simple average. Our aggregation module synthesizes the multi-level evidence using a ``candidate majority voting within probed paths'' mechanism to resolve cross-level divergences. Specifically, votes from different probed levels (e.g., an \textit{above} vote at L3 and a \textit{below} vote at L4) cast support for their corresponding candidate levels within the actually explored trajectory. The level accumulating the most votes is selected as the overall quality grade; in the event of a tie, the lowest candidate level is conservatively chosen. Concurrently, we enforce a strict \textit{Structural Ceiling}: if the structure dimension is evaluated as \textit{below} at any probed level $k$, the structural ceiling is strictly capped at $k-1$. The final grade is conservatively computed as $\text{Final Level} = \min(\text{Structural Ceiling}, \text{Overall Quality})$. This strategy effectively resolves conflicts between different match results, guaranteeing that the final aesthetic score strictly adheres to the baseline of structural integrity.

\subsection{Visual Text Rendering}

One of the core challenges in visual text rendering lies in accurately generating a large number of characters while maintaining correct spelling and complex spatial typography. Previous evaluation metrics for visual text rendering predominantly rely on Edit Distance to measure the accuracy of generated text. Nevertheless, due to the complex typographic structures present in image text, traditional methods are often only applicable to simple scenarios and heavily depend on the recognition accuracy of Optical Character Recognition (OCR) models. Recently, some approaches~\cite{zhu2026textpecker} have attempted to substitute OCR models with Vision-Language Models for text recognition. Nevertheless, constrained by the inherent hallucination phenomenon of Vision-Language Models, these methods are highly prone to ``imagining'' incorrect text as correct, leading to artificially inflated accuracy scores.

To comprehensively and objectively evaluate the accuracy, coherence, and readability of models in text generation, we integrate and establish five core text evaluation metrics: Edit Similarity ($\text{Sim}_{\text{edit}}$) and Sentence Accuracy ($\text{Acc}_{\text{sen}}$) based on OCR models, alongside Generalized Normalized Edit Distance ($\text{GNED}$), Quality/Readability ($\text{Score}_{\text{qua}}$), and Character-level Precision/Recall/F1 scores ($\text{Char}_{\text{P/R/F1}}$) based on TextPecker~\cite{zhu2026textpecker}. Notably, TextPecker effectively mitigates the hallucination problem of Vision-Language Models during text recognition tasks by fine-tuning them on a specially constructed dataset. Furthermore, we introduce a universal enhancement across all evaluation metrics: by pre-matching the minimum $\mathrm{NED}$ cost to concatenate multi-segment text content, this modification significantly bolsters the evaluation robustness in scenarios involving multi-segment texts or complex typography. A detailed analysis of these five metrics is presented as follows:

\textbf{Edit Similarity ($\text{Sim}_{\text{edit}}$).} This metric measures the global similarity between the Ground Truth ($\mathrm{GT}$) and the Prediction ($\mathrm{Pred}$) through a minimum concatenation cost strategy. Specifically, this method first concatenates all text segments contained in the $\mathrm{GT}$ and $\mathrm{Pred}$ in the optimal order, and subsequently calculates the Normalized Edit Distance ($\mathrm{NED}$) between the two concatenated full strings, calculated as Eq.~\ref{eq:sim_edit}:
\begin{equation}
\text{Sim}_{\text{edit}} = 1 - \mathrm{NED}(\mathrm{GT}_{\mathrm{joined}}, \mathrm{Pred}_{\mathrm{joined}}).
\label{eq:sim_edit}
\end{equation}
The primary advantage of $\text{Sim}_{\text{edit}}$ is its inter-segment minimum cost concatenation strategy, which effectively overcomes line-break misalignments caused by typography in text-to-image tasks and the issue of OCR tools mistakenly splitting long sentences into multiple segments, while preserving the text sequence constraints within segments. Nonetheless, this metric only outputs a global similarity scalar and cannot distinguish at a multi-dimensional level whether the model produced omissions, typos, or severe generation hallucinations, which is detrimental to deep error attribution.

\textbf{Sentence Accuracy ($\text{Acc}_{\text{sen}}$).} To measure the model's capability for ``perfect generation,'' $\text{Acc}_{\text{sen}}$ adopts a macroscopic perspective of segment matching. For $N$ text segments in the $\mathrm{GT}$ and $K$ text segments in the $\mathrm{Pred}$, this metric uses $\mathrm{NED}$ as the cost matrix, employs maximum weight matching in bipartite graphs (the Hungarian algorithm) for global alignment, and calculates the proportion of perfectly matched segments (Eq.~\ref{eq:acc_sen}):
\begin{equation}
\text{Acc}_{\text{sen}} = \frac{\text{Number of Perfectly Matched Segments}}{\text{Total Number of Segments in } \mathrm{GT}}.
\label{eq:acc_sen}
\end{equation}
By introducing Hungarian matching, this metric successfully resolves matching difficulties caused by unequal segment numbers and different spatial arrangements between $\mathrm{GT}$ and $\mathrm{Pred}$. However, a notable limitation is its stringent evaluation criterion. It is highly sensitive to abnormal text splitting caused by OCR systems; even text that is visually correct but slightly different in typographic splitting will be judged as a generation error.

\textbf{Quality / Readability ($\text{Score}_{\text{qua}}$).} Compared to strict semantic matching, the $\text{Score}_{\text{qua}}$ metric focuses on the visual rendering quality of the generated text itself, i.e., whether the text is clear and legible. In practical calculations, OCR tools are typically used to detect unrecognizable abnormal regions and occupy them with specific tokens (e.g., \texttt{<\#>}), followed by calculating the ratio of valid characters to the total generated characters, as shown in Eq.~\ref{eq:score_qua}:
\begin{equation}
\text{Score}_{\text{qua}} = 1 - \frac{\text{Number of Abnormal Characters}}{\text{Total Characters in Predicted Text}}.
\label{eq:score_qua}
\end{equation}
This metric intuitively reflects the model's clarity performance at the visual level, eliminating interference from whether the text semantics align with the target. Nevertheless, its limitation is that it largely ignores the semantic correctness of the text content. A model that generates highly legible text but deviates significantly from the $\mathrm{GT}$ semantics (e.g., severe hallucination) might still obtain a high $\text{Score}_{\text{qua}}$ score; thus, it cannot be used independently to evaluate the model's text compliance capability.

\textbf{Generalized Normalized Edit Distance ($\text{GNED}$).} The $\text{GNED}$ metric emphasizes difference statistics at the character set level. It first tokenizes the $\mathrm{GT}$ and $\mathrm{Pred}$, then uses the Hungarian algorithm to find the minimum-cost character mapping, and finally counts the number of unmatched tokens via Eq.~\ref{eq:gned}:
\begin{equation}
\text{GNED} = 1 - \frac{\text{Total Number of Unmatched Tokens}}{\max(|\mathcal{T}_{\mathrm{GT}}|, |\mathcal{T}_{\mathrm{Pred}}|)}.
\label{eq:gned}
\end{equation}
Here, the number of unmatched tokens is equivalent to the sum of differences caused by extra, missing, or incorrect characters. The advantage of $\text{GNED}$ is its ability to capture all microscopic character differences that are unaligned at both ends, providing an overview of character-level accuracy. Nonetheless, because it relaxes the sequential order constraints between texts during matching, it may occasionally lead the system to overestimate the model's true accuracy, and similarly lacks a detailed classification of different error types.

\textbf{Character-level Precision, Recall, and F1 Score ($\text{Char}_{\text{P/R/F1}}$).} To compensate for the shortcomings of the aforementioned metrics in error attribution, $\text{Char}_{\text{P/R/F1}}$ combines segment-level Hungarian matching with fine-grained character-level edit alignment. In the aligned long sentence sequences, True Positives ($\mathrm{TP}$), False Positives ($\mathrm{FP}$), and False Negatives ($\mathrm{FN}$) are counted based on character-level Substitution, Deletion, and Insertion operations, defined in Eq.~\ref{eq:char_metrics}:
\begin{equation}
\text{Precision} = \frac{\mathrm{TP}}{\mathrm{TP} + \mathrm{FP}}, \quad \text{Recall} = \frac{\mathrm{TP}}{\mathrm{TP} + \mathrm{FN}}, \quad \text{F1} = \frac{2 \times \text{Precision} \times \text{Recall}}{\text{Precision} + \text{Recall}}.
\label{eq:char_metrics}
\end{equation}
This system possesses excellent model diagnostic capabilities: a low $\text{Precision}$ implies excessive false detections (high $\mathrm{FP}$), directly pointing to hallucinations or typos; a low $\text{Recall}$ indicates severe character omissions (high $\mathrm{FN}$). This method simultaneously accounts for the macroscopic arrangement order of text and microscopic character accuracy. However, its main limitation is the extremely high computational complexity of the algorithm; in evaluation scenarios involving multi-segment, long texts, complex matching combinations can easily trigger combinatorial explosion problems.

\section{Experiments}

\subsection{Validation of Evaluation Metrics}

\subsubsection{Human Ground Truth Construction and Correlation Analysis}
To rigorously validate the effectiveness of our proposed evaluation metrics, we establish a robust human preference baseline via pairwise comparative analysis. Specifically, we randomly sample two images generated by different models for each of the 2,000 prompts in the WeGenBench-General subset, forming a comprehensive set of image pairs. Human experts are then tasked with conducting side-by-side evaluations. It is important to note that the annotations for aesthetic quality and semantic alignment are conducted independently to prevent cross-contamination of judgments. For \textit{Aesthetic Quality}, we enforce a strict forced-choice paradigm, requiring evaluators to definitively assign a ``Win'' or ``Loss'' to capture subtle perceptual preferences. Conversely, for \textit{Semantic Alignment}, we introduce an additional ``Tie'' option, acknowledging the practical scenario where both generated images may equally satisfy or violate the conditioning prompt. These independent, human-annotated outcomes serve as the ground-truth preference labels for our validation.

Subsequently, we apply both our proposed metrics and existing baseline methods to automatically evaluate the identical image pairs. Given the inclusion of ``Tie'' cases in the alignment assessment and the discrete ordinal nature of our proposed scoring methods (e.g., 1-5 or 1-10 scales), relying solely on traditional metrics like overall win rate, strict 3-class accuracy, or Cohen's Kappa introduces severe statistical bias. These conventional metrics implicitly favor continuous scoring distributions and unfairly penalize discrete methods that correctly and frequently predict ties, often treating such valid tie predictions as noise.

To address this, we formulate the validation across both Pair-level and Model-level dimensions using a suite of metrics specifically designed to be robust to discrete outputs and fair to tie predictions. At the Pair-level, we utilize \textit{Pair AUC} and \textit{Decisive Accuracy}. \textit{Pair AUC} is calculated exclusively on the subset of pairs where both human evaluators and the automated method make decisive, non-tie judgments. This approach prevents the metric from being artificially deflated by a method's inherent tie-prediction rate, offering an unbiased measure of pure ranking quality. Similarly, \textit{Decisive Accuracy} measures the alignment hit rate only when the automated method explicitly declares a clear winner, isolating its true discriminative accuracy from its threshold for declaring a tie.

At the Model-level, rather than relying on pairwise win-rate aggregation---which inherently discards the magnitude of score differences---we aggregate the per-image mean scores across the entire dataset to rank the 16 evaluated models. This mean-score aggregation fully utilizes the ordinal score information provided by the evaluators. Based on these model rankings, we compute the \textit{Spearman $\rho$} and \textit{Kendall $\tau$-b} rank correlations, alongside the Mean Absolute Rank Difference (\textit{MARD}) against the human leaderboard. These model-level metrics provide a holistic and highly stable reflection of how well an automated metric aligns with human consensus on overall model capabilities.

\begin{table}[h]
  \centering
  \caption{Validation of evaluation metrics against human judgments. We evaluate automated metrics across both Pair-level and Model-level dimensions. Pair AUC (AUC drop both ties) and Decisive Acc. measure the pair-level ranking quality, while Model $\rho$, Model $\tau$-b, and Model MARD measure the consistency with human model leaderboards. Our proposed metrics demonstrate superior correlation and lower rank deviation compared to existing baselines. Best results in each category are highlighted in \textbf{bold}.}
  \label{tab:metric_validation}
  \resizebox{0.98\linewidth}{!}{%
  \renewcommand{\arraystretch}{1.2}
  \begin{tabular}{@{}l ccccc @{}}
    \toprule
    \textbf{Method} & \textbf{Pair AUC $\uparrow$} & \textbf{Decisive Acc. $\uparrow$} & \textbf{Model $\rho$ $\uparrow$} & \textbf{Model $\tau$-b $\uparrow$} & \textbf{Model MARD $\downarrow$} \\
    \midrule
    \multicolumn{6}{l}{\textbf{Aesthetics \& Preference}} \\
    \midrule
    ImageReward~\cite{xu2023imagereward} & 0.51 & 0.52 & 0.05 & -0.01 & 5.25 \\
    HPSv2~\cite{wu2023human}             & 0.54 & 0.53 & \underline{0.24} & \underline{0.23} & 4.88 \\
    HPSv3~\cite{ma2025hpsv3}             & \underline{0.60} & \underline{0.57} & \underline{0.24} & 0.19 & \underline{4.75} \\
    \rowcolor{gray!10}
    \textbf{Anchor-based Match (Ours)}   & \textbf{0.63} & \textbf{0.61} & \textbf{0.83} & \textbf{0.68} & \textbf{2.06} \\
    \midrule
    \multicolumn{6}{l}{\textbf{Semantic Alignments}} \\
    \midrule
    FG-CLIP 2~\cite{xie2025fg}                              & 0.63 & 0.58 & 0.58 & 0.44 & 2.81 \\
    VQAScore~\cite{lin2024evaluating}    & 0.70 & 0.65 & \underline{0.87} & \underline{0.71} & 1.88 \\
    \rowcolor{gray!10}
    \textbf{Checklist-based QA (Ours)}   & \underline{0.73} & \underline{0.66} & \underline{0.87} & \underline{0.71} & \underline{1.75} \\
    \rowcolor{gray!10}
    \textbf{COT-based Deduction (Ours)}  & \textbf{0.75} & \textbf{0.71} & \textbf{0.91} & \textbf{0.78} & \textbf{1.25} \\
    \bottomrule
  \end{tabular}%
  }
\end{table}

As demonstrated in Tab.~\ref{tab:metric_validation}, our proposed methods exhibit a decisive advantage over existing baselines across both aesthetic and alignment tasks. In the Aesthetic evaluation, our \textbf{Anchor-based Match} achieves a Model Spearman $\rho$ of $0.83$, vastly outperforming HPSv3 ($0.24$) and HPSv2 ($0.24$), while reducing the average rank deviation (MARD) to just $2.06$ positions. In the Semantic Alignment evaluation, our \textbf{COT-based Deduction} consistently ranks first across all five metrics, achieving an exceptional Model Spearman $\rho$ of $0.91$ and a Pair AUC of $0.75$. Our \textbf{Checklist-based QA} metric also demonstrates highly competitive performance, securing the second-best results in Pair AUC ($0.73$) and MARD ($1.75$). These results confirm that our multi-dimensional evaluation framework not only provides multi-dimensional interpretability but also aligns significantly better with human cognitive judgments than current state-of-the-art automated evaluators.

\subsubsection{Qualitative Validation of Metrics}

\begin{figure}[h]
  \centering
  \includegraphics[width=\linewidth]{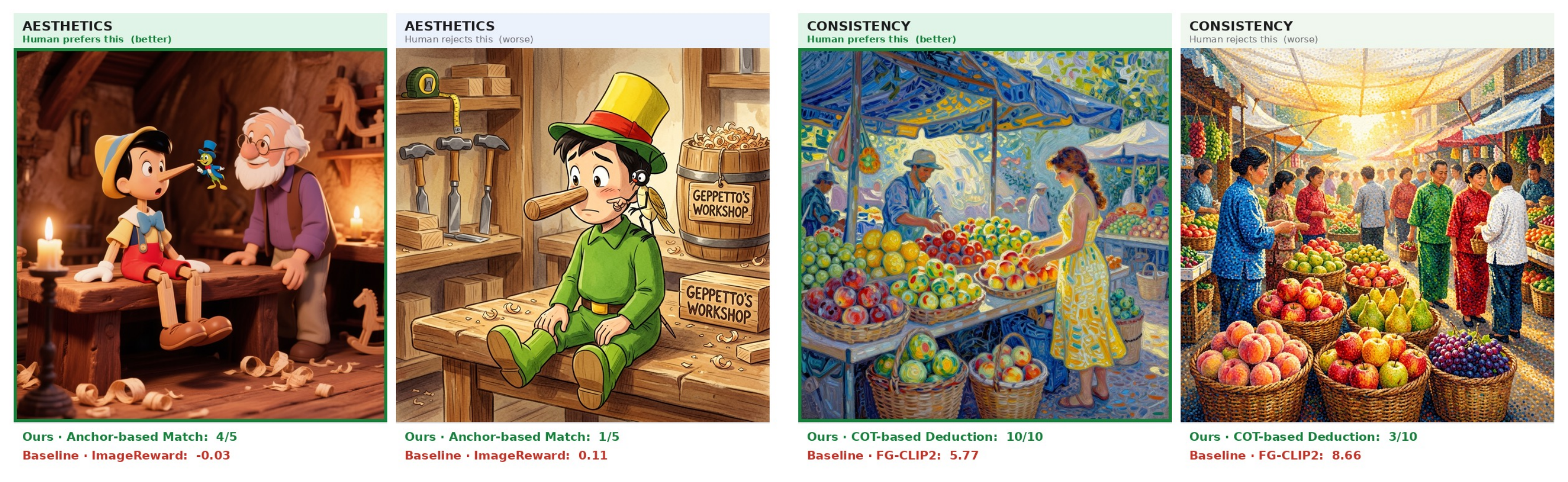} 
  \caption{\textbf{Qualitative validation of evaluation metrics.} The 1$\times$4 grid illustrates cases where baseline metrics fail to align with human perception. \textbf{Right (Alignment):} Baselines like CLIPScore~\cite{hessel2021clipscore} act as shallow feature matchers, giving high scores despite severe attribute leakage or binding errors, whereas our COT and QA metrics accurately penalize these localized flaws. \textbf{Left (Aesthetics):} Baselines like ImageReward~\cite{xu2023imagereward} are deceived by vibrant styles and completely ignore severe structural deformities, while our Anchor-based Match correctly applies a structural ceiling to downgrade the score.}
  \label{fig:metric_qualitative}
\end{figure}

To intuitively demonstrate why our proposed metrics achieve higher correlation with human perception than existing baselines, we provide qualitative comparisons in Fig.~\ref{fig:metric_qualitative}. 

In terms of \textit{Semantic Alignment} (Fig.~\ref{fig:metric_qualitative}, left), traditional cross-modal metrics like CLIPScore often behave as shallow feature matchers. They heavily reward the mere presence of requested objects and colors, while completely failing to comprehend complex spatial relationships, precise counting, or attribute binding. For instance, when a model generates an image with severe attribute leakage (e.g., swapping the colors of two subjects), CLIPScore still assigns a misleadingly high score because all semantic keywords are visually present. In contrast, our COT-based Deduction and Checklist-based QA systematically deconstruct the prompt and perform rigorous logical reasoning, successfully penalizing these localized binding errors that baselines overlook. 

Regarding \textit{Aesthetic Quality} (Fig.~\ref{fig:metric_qualitative}, right), existing scoring models like ImageReward frequently suffer from ``first-glance bias.'' They tend to assign high scores to images with vibrant colors, high contrast, or highly stylized textures, remaining entirely blind to severe structural deformities (such as distorted limbs, asymmetrical facial features, or physically impossible geometries). Our Anchor-based Match Grading overcomes this critical flaw by introducing a strict ``Structural Ceiling'' mechanism. By explicitly decoupling structural correctness from stylistic appeal, our metric ensures that structural integrity acts as a fundamental prerequisite for high aesthetic ratings, perfectly mirroring the rigorous, multi-dimensional standards of human experts.

\subsection{Benchmarking Existing Models}
\label{sec:benchmarking}

To comprehensively evaluate the capabilities of current state-of-the-art text-to-image generation models, we conduct extensive benchmarking on WeGenBench. Our evaluation encompasses a diverse array of models, including both leading open-source diffusion models (e.g., FLUX.2-dev~\cite{black-forest-labs-flux-2}, HunyuanImage-3.0~\cite{cao2025hunyuanimage}, Qwen-Image~\cite{wu2025qwen}, Z-Image~\cite{cai2025z}, GLM-Image~\cite{glm-image}, ERNIE-Image-Turbo~\cite{ernie-image}, SenseNova-U1-8B-MoT~\cite{diao2026sensenova}, HiDream-O1-Image~\cite{hidreamolimage}) and cutting-edge closed-source commercial systems (e.g., GPT-Image-2~\cite{gpt-image-2}, Seedream-4.5~\cite{seedream4_5}, Nano-banana-2~\cite{nanobanana2}). To ensure a fair and reproducible evaluation, we strictly adhere to the default configurations and hyperparameters recommended by the official implementations for all open-source models, without any additional manual tuning. For the automated evaluation engines, we employ the Qwen3.5-397B-A17B-FP8 model for both the Checklist-based QA verification and the Anchor-based Match aesthetic grading. The unified prompt templates and detailed configurations utilized for these two VLM evaluators are provided in Appendix~\ref{sec:appendix_prompts}. 

It is worth noting that due to stringent safety filters and copyright restrictions inherent in commercial APIs, the closed-source models failed to generate valid outputs for a small fraction of the test prompts. Specifically, out of the 2,000 General prompts, GPT-Image-2, Seedream-4.5, and Nano-banana-2 successfully generated 1,961, 1,993, and 1,964 images, respectively. Similarly, for the 2,000 Text Rendering prompts, GPT-Image-2, Seedream-4.5, and Nano-banana-2 successfully generated 1,982, 1,987, and 1,984 images, respectively. The quantitative results across the three core dimensions are calculated strictly based on these successfully generated samples. Specifically, the Semantic Alignment (QA, COT) and Aesthetic Quality (AM) scores in Tab.~\ref{tab:main_results} are evaluated exclusively on the 2,000-prompt General subset, whereas the five Visual Text Rendering metrics ($\text{Sim}_{\text{edit}}$, $\text{Acc}_{\text{sen}}$, $\text{Score}_{\text{qua}}$, $\text{GNED}$, $\text{Char}_{\text{F1}}$) are evaluated exclusively on the 2,000-prompt Text subset. For missing samples caused by API rejections, we exclude them from the denominator during the average score aggregation to prevent unfair penalization of the models' generative capabilities due to safety policies.

\subsubsection{Quantitative Analysis}

To provide a comprehensive and interpretable evaluation, we introduce the specific metrics utilized across our three primary dimensions. 
For \textbf{Semantic Alignment}, we employ two metrics: \textbf{Checklist-based QA (QA)}, which calculates the pass rate of fine-grained visual questions to reflect the model's basic semantic adherence, and \textbf{COT-based Deduction (COT)}, which evaluates complex logical alignment using Chain-of-Thought reasoning to reflect deep semantic understanding and compositionality. 
For \textbf{Aesthetic Quality}, we use \textbf{Anchor-based Match (AM)}, which scores images from 0 to 5 based on structural integrity and stylistic appeal, reflecting the overall visual generation quality. 
For \textbf{Visual Text Rendering}, we utilize five metrics: \textbf{Edit Similarity ($\text{Sim}_{\text{edit}}$)} (normalized Levenshtein distance) and \textbf{Sentence Accuracy ($\text{Acc}_{\text{sen}}$)} (exact match rate) to reflect character-level spelling accuracy; \textbf{Quality Score ($\text{Score}_{\text{qua}}$)} to assess typographic aesthetics and readability; \textbf{GNED} (Generalized Normalized Edit Distance) to measure robustness against spatial layout variations; and \textbf{Character F1 ($\text{Char}_{\text{F1}}$)} to reflect the overall balance of text generation without hallucination or omission.

\begin{table}[h]
  \centering
  \caption{Comprehensive benchmarking results of state-of-the-art text-to-image generation models on WeGenBench. The evaluation is categorized into three primary dimensions: Aesthetics, Alignments, and Visual Text Rendering. AM: Anchor-based Match Grading, QA: Checklist-based QA, COT: COT-based Deduction.}
  \label{tab:main_results}
  \resizebox{0.98\linewidth}{!}{%
  \renewcommand{\arraystretch}{1.2}
  \setlength{\tabcolsep}{2pt}
  \begin{tabular}{@{}l cc c ccccc @{}}
    \toprule
    & \multicolumn{2}{c}{\textbf{Alignments}} & \textbf{Aesthetics} & \multicolumn{5}{c}{\textbf{Visual Text Rendering}} \\
    \cmidrule(lr){2-3} \cmidrule(lr){4-4} \cmidrule(lr){5-9}
    \textbf{Models} & QA $\uparrow$ & COT $\uparrow$ & AM $\uparrow$ & $\text{Sim}_{\text{edit}}$ $\uparrow$ & $\text{Acc}_{\text{sen}}$ $\uparrow$ & $\text{Score}_{\text{qua}}$ $\uparrow$ & $\text{GNED}$ $\uparrow$ & $\text{Char}_{\text{F1}}$ $\uparrow$ \\
    \midrule
    \multicolumn{9}{l}{\textcolor{black!70}{\textit{Open-Source Models}}} \\
    \midrule
    GLM-Image~\cite{glm-image} & 0.86 & 6.23 & 2.58 & \underline{0.73} & 0.56 & \underline{0.92} & \underline{0.74} & \underline{0.78} \\
    FLUX.2-dev~\cite{black-forest-labs-flux-2} & 0.93 & 7.91 & 2.69 & 0.51 & 0.38 & 0.76 & 0.55 & 0.58 \\
    Qwen-Image~\cite{wu2025qwen} & 0.91 & 7.44 & 2.56 & 0.45 & 0.33 & 0.61 & 0.43 & 0.46 \\
    Qwen-Image-2512~\cite{wu2025qwen} & 0.90 & 7.43 & 2.70 & 0.62 & 0.43 & 0.72 & 0.57 & 0.59 \\
    Z-Image~\cite{cai2025z} & 0.92 & 7.65 & 2.44 & 0.50 & 0.53 & 0.78 & 0.52 & 0.55 \\
    Z-Image-Turbo~\cite{cai2025z} & 0.90 & 7.26 & 2.67 & 0.67 & 0.50 & 0.87 & 0.69 & 0.73 \\
    HunyuanImage-3.0-Inst. (image)~\cite{cao2025hunyuanimage} & 0.95 & 8.26 & 2.65 & 0.63 & 0.56 & 0.88 & 0.66 & 0.69 \\
    HunyuanImage-3.0-Inst. (think\_re.)~\cite{cao2025hunyuanimage} & 0.95 & 8.32 & 2.74 & 0.56 & 0.56 & 0.90 & 0.57 & 0.63 \\
    HunyuanImage-3.0-Inst.-D (image)~\cite{cao2025hunyuanimage} & 0.93 & 8.00 & 2.62 & 0.66 & 0.50 & 0.85 & 0.67 & 0.70 \\
    HunyuanImage-3.0-Inst.-D (think\_re.)~\cite{cao2025hunyuanimage} & 0.94 & 8.08 & 2.70 & 0.57 & 0.54 & 0.89 & 0.57 & 0.63 \\
    Ernie-Image-Turbo (pe)~\cite{ernie-image} & 0.92 & 7.72 & 2.59 & 0.60 & 0.56 & 0.90 & 0.61 & 0.65 \\
    Ernie-Image-Turbo (no-PE)~\cite{ernie-image} & 0.92 & 7.87 & 2.65 & 0.70 & 0.56 & 0.89 & 0.71 & 0.75 \\
    SenseNova-U1-8B-MoT~\cite{diao2026sensenova} & 0.92 & 7.72 & 2.38 & 0.68 & 0.52 & 0.90 & 0.71 & 0.74 \\
    SenseNova-U1-8B-MoT-think~\cite{diao2026sensenova} & 0.93 & 7.81 & 2.41 & 0.68 & 0.51 & 0.88 & 0.70 & 0.72 \\
    HiDream-O1~\cite{hidreamolimage} & 0.87 & 6.11 & 2.30 & 0.49 & 0.54 & 0.78 & 0.56 & 0.57 \\
    \midrule
    \multicolumn{9}{l}{\textcolor{black!70}{\textit{Closed-Source Models}}} \\
    \midrule
    GPT-Image-2~\cite{gpt-image-2} & \textbf{0.97} & \textbf{8.95} & \textbf{2.84} & 0.50 & \underline{0.57} & 0.88 & 0.52 & 0.56 \\
    Seedream-4.5~\cite{seedream4_5} & 0.96 & \underline{8.74} & \underline{2.81} & \textbf{0.74} & \textbf{0.59} & \textbf{0.93} & \textbf{0.76} & \textbf{0.79} \\
    Nano-banana-2~\cite{nanobanana2} & \underline{0.97} & 8.71 & 2.77 & 0.38 & 0.55 & 0.86 & 0.45 & 0.49 \\
    \bottomrule
  \end{tabular}%
  }
  
  \vspace{2pt}
\raggedright \scriptsize $^*$ The Automatic Prompt Enhancement (APE) feature may introduce unprompted text elements, leading to a decrease in strict text rendering metrics. This phenomenon is particularly prevalent in closed-source commercial models (i.e., GPT-Image-2, Seedream-4.5, and Nano-banana-2) and open-source models operating under reasoning-enhanced configurations (i.e., HunyuanImage-3.0-Inst. (think\_re.), HunyuanImage-3.0-Inst.-D (think\_re.), Ernie-Image-Turbo (pe), and SenseNova-U1-8B-MoT-think).
\end{table}

As shown in Tab.~\ref{tab:main_results} and further supported by our multi-dimensional tagging system, the benchmarking results reveal several universal behavioral patterns across all evaluated T2I models, as well as distinct domain-specific characteristics for individual models.

\begin{figure}[h]
  \centering
  \includegraphics[width=\linewidth]{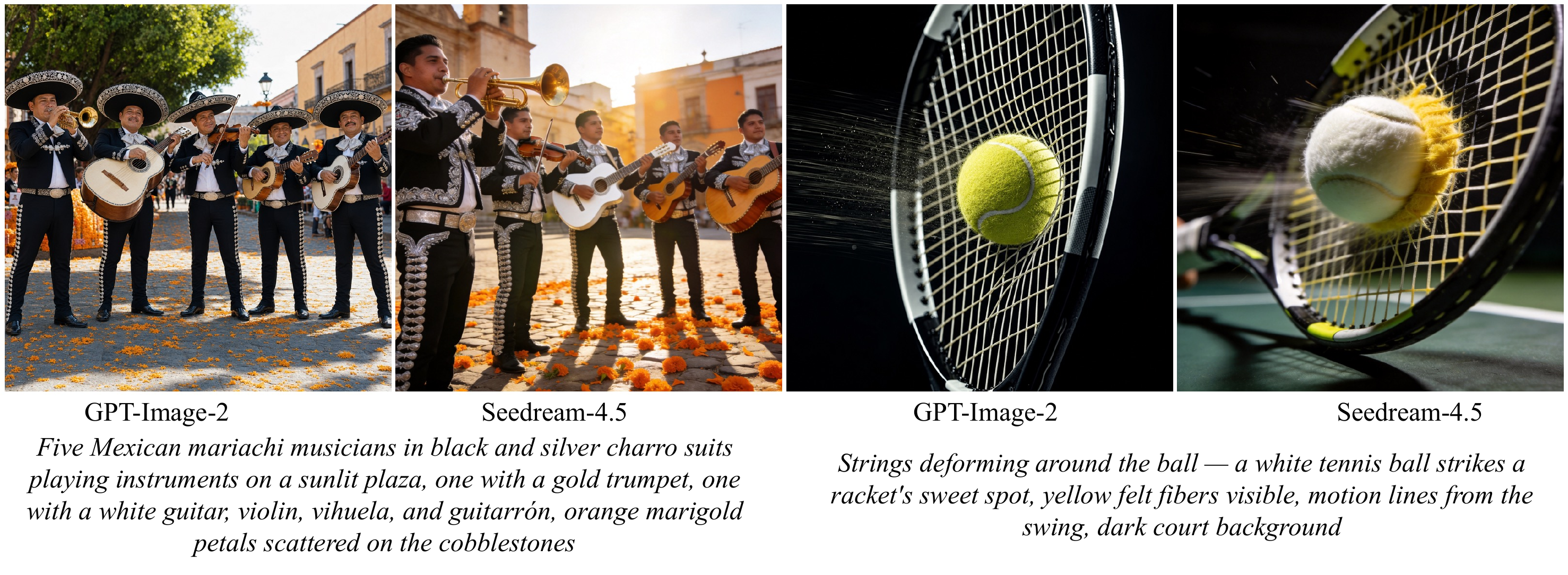}
  \caption{\textbf{Universal bottlenecks shared across SOTA T2I models.} \textbf{Left:} A prompt requesting five Mexican musicians each holding a distinct instrument illustrates a multi-subject attribute-binding failure: even GPT-Image-2 and Seedream-4.5 misassign or drop the per-subject instruments when the cardinality and rarity of attributes increase. \textbf{Right:} A prompt that explicitly requires a \emph{white} tennis ball exposes a strong visual prior overriding the explicit instruction; GPT-Image-2 still produces the statistically dominant yellow-green ball, showing that deeply ingrained world knowledge can override the user's explicit prompt.}
  \label{fig:bottlenecks}
\end{figure}

\textbf{Universal Laws and Bottlenecks.} First, we observe a striking orthogonality between macroscopic semantic comprehension and microscopic text rendering accuracy. While models may excel at rendering complex scenes, this does not naturally translate into precise spelling, and vice versa. In the domain of Visual Text Rendering, the results expose a universal bottleneck. While several models achieve exceptionally high Quality/Readability scores ($\text{Score}_{\text{qua}} > 0.90$), their performance drops significantly on strict semantic matching metrics (e.g., $\text{Acc}_{\text{sen}}$), indicating that models excel at generating visually clear, text-like features but struggle with character-level spelling accuracy. A particularly revealing example in this regard is the rendering of English UI interfaces, where even the top-performing models fail to surpass a character-level F1 of $0.55$, suggesting that small-font, densely packed Latin typography in compact layouts remains an unsolved challenge across the entire field. Second, regardless of a model's overall capability, prompt length and complexity remain universal bottlenecks (Fig.~\ref{fig:bottlenecks}). All evaluated models exhibit a significant performance surge when processing short prompts (often scoring $+1.0$ to $+1.7$ higher than their respective averages), but their performance universally degrades when confronted with tags requiring ``multi-subject,'' ``complex action,'' ``complex interaction,'' or ``precise quantity.'' Moreover, ``negation'' and ``counting'' remain the most vulnerable capabilities universally; even top-tier models struggle to consistently follow ``do not draw'' instructions or accurately render specific quantities of objects. Finally, our bilingual evaluation reveals a persistent asymmetric performance in cross-lingual generation. Across all models, semantic alignment scores for Chinese prompts are consistently lower than those for English prompts (with average drops ranging from $0.4$ to $5.0$ points). Interestingly, in the specific domain of visual text rendering, most models perform slightly better at generating Chinese characters than English words, likely due to the shorter average string lengths and more standardized typographic layouts typical of Chinese text prompts in our dataset. It is also important to note that our evaluation metrics for text rendering and semantic alignment inherently rely on the foundational capabilities of the underlying VLMs and OCR models. Consequently, the absolute scores may be partially influenced by the recognition limits of these evaluator models, though the relative performance rankings remain highly consistent.

\begin{table}[h]
  \centering
  \small
  \renewcommand{\arraystretch}{1.2}
  \caption{\textbf{Aesthetic Level Distribution.} The proportion of generated images falling into each aesthetic quality level (L0 to L5) as evaluated by our Anchor-based Match Grading scheme. Higher L4 and L5 percentages indicate stronger aesthetic capabilities.}
  \label{tab:aes_distribution}
  \resizebox{\linewidth}{!}{%
  \begin{tabular}{@{}l c cccccc @{}}
    \toprule
    \textbf{Models} & \textbf{Mean Score $\uparrow$} & \textbf{L0 (\%) $\downarrow$} & \textbf{L1 (\%) $\downarrow$} & \textbf{L2 (\%) $\downarrow$} & \textbf{L3 (\%) $\uparrow$} & \textbf{L4 (\%) $\uparrow$} & \textbf{L5 (\%) $\uparrow$} \\
    \midrule
    \multicolumn{8}{l}{\textcolor{black!70}{\textit{Open-Source Models}}} \\
    \midrule
    GLM-Image~\cite{glm-image} & 2.58 & 3.30 & 7.20 & 26.25 & 54.35 & 8.80 & 0.10 \\
    FLUX.2-dev~\cite{black-forest-labs-flux-2} & 2.69 & 2.70 & \textbf{4.60} & 23.00 & 60.70 & 8.90 & 0.10 \\
    Qwen-Image~\cite{wu2025qwen} & 2.56 & 4.45 & 6.60 & 26.50 & 53.95 & 8.35 & 0.15 \\
    Qwen-Image-2512~\cite{wu2025qwen} & \underline{2.70} & 2.60 & 6.95 & 24.55 & 49.95 & \textbf{15.65} & \underline{0.30} \\
    Z-Image~\cite{cai2025z} & 2.44 & 6.15 & 8.55 & 27.25 & 51.75 & 6.15 & 0.15 \\
    Z-Image-Turbo~\cite{cai2025z} & 2.67 & 1.95 & 6.00 & \textbf{22.55} & 62.20 & 7.10 & 0.20 \\
    HunyuanImage-3.0-Inst. (image)~\cite{cao2025hunyuanimage} & 2.65 & 4.15 & 5.05 & 23.60 & 55.85 & 11.15 & 0.20 \\
    HunyuanImage-3.0-Inst. (think\_re.)~\cite{cao2025hunyuanimage} & \textbf{2.74} & \textbf{1.40} & \underline{4.75} & 23.65 & 59.25 & 10.60 & \textbf{0.35} \\
    HunyuanImage-3.0-Inst.-D (image)~\cite{cao2025hunyuanimage} & 2.62 & 4.15 & 6.75 & 23.60 & 54.05 & \underline{11.25} & 0.20 \\
    HunyuanImage-3.0-Inst.-D (think\_re.)~\cite{cao2025hunyuanimage} & 2.70 & \underline{1.85} & 5.55 & 24.25 & 57.80 & 10.40 & 0.15 \\
    Ernie-Image-Turbo (pe)~\cite{ernie-image} & 2.59 & 3.15 & 6.65 & 28.30 & 52.45 & 9.10 & \textbf{0.35} \\
    Ernie-Image-Turbo (no-PE)~\cite{ernie-image} & 2.65 & 2.65 & 6.60 & \underline{22.65} & 59.25 & 8.60 & 0.25 \\
    SenseNova-U1-8B-MoT~\cite{diao2026sensenova} & 2.38 & 4.70 & 12.05 & 32.45 & 42.60 & 7.90 & \underline{0.30} \\
    SenseNova-U1-8B-MoT-think~\cite{diao2026sensenova} & 2.41 & 3.75 & 13.25 & 31.45 & \underline{41.75} & 9.75 & 0.05 \\
    HiDream-O1~\cite{hidreamolimage} & 2.30 & 5.90 & 13.20 & 32.40 & \textbf{41.75} & 6.70 & 0.05 \\
    \midrule
    \multicolumn{8}{l}{\textcolor{black!70}{\textit{Closed-Source Models}}} \\
    \midrule
    GPT-Image-2~\cite{gpt-image-2} & \textbf{2.84} & \underline{1.02} & \textbf{3.21} & \underline{22.23} & \underline{57.73} & \textbf{15.45} & \underline{0.36} \\
    Seedream-4.5~\cite{seedream4_5} & \underline{2.81} & 1.35 & \underline{3.71} & \textbf{20.97} & 60.91 & \underline{12.69} & 0.35 \\
    Nano-banana-2~\cite{nanobanana2} & 2.77 & \textbf{1.02} & 3.87 & 22.61 & 62.78 & 9.22 & \textbf{0.51} \\
    \bottomrule
  \end{tabular}%
  }
\end{table}

As detailed in Tab.~\ref{tab:aes_distribution}, we further decompose the overall aesthetic scores into a fine-grained level distribution (L0 to L5) to provide a more nuanced understanding of each model's generative capabilities. In our Anchor-based Match Grading scheme, L0 and L1 represent severe structural failures or meaningless noise, L2 and L3 indicate mediocre to acceptable images with minor flaws, and L4 and L5 represent top-tier images with flawless structure and high aesthetic appeal. This distribution reveals that while mean scores may appear similar, the underlying quality consistency varies significantly. For instance, GPT-Image-2 and Seedream-4.5 not only achieve the highest mean scores (2.84 and 2.81, respectively) but also dominate the top-tier quality brackets, with $15.81\%$ and $13.04\%$ of their generated images reaching L4 or L5. In contrast, models with lower mean scores, such as GLM-Image (2.58), only achieve $8.90\%$ in the L4 and L5 brackets combined. Notably, Qwen-Image-2512 exhibits a unique ``high-ceiling'' characteristic among open-source models; although its mean score is slightly lower than the top commercial systems, it produces the highest proportion of L4 images ($15.65\%$) among all open-source models, albeit accompanied by a slightly higher variance in the lower tiers. Conversely, models like \textbf{Z-Image} show a heavy concentration in the L2 and L3 brackets, indicating a baseline capability to produce acceptable images but a struggle to consistently achieve exceptional aesthetic appeal. This level-wise breakdown underscores that a model's true aesthetic proficiency is defined not just by its average performance, but by its ability to reliably push the upper boundaries of visual quality while minimizing severe structural failures (L0 and L1).

\begin{table*}[h]
  \centering
  \small
  \renewcommand{\arraystretch}{1.2}
  \caption{\textbf{Detailed Visual Text Rendering Analysis.} Character-level Precision (P), Recall (R), and F1 scores evaluated using both our VLM-based metric (TextPecker) and traditional OCR. Precision reflects the absence of hallucinations/typos, while Recall reflects the completeness of the rendered text.}
  \label{tab:text_pr_analysis}
  \resizebox{\linewidth}{!}{%
  \begin{tabular}{@{}l | ccc | ccc @{}}
    \toprule
    & \multicolumn{3}{c|}{\textbf{VLM-based (TextPecker)}} & \multicolumn{3}{c}{\textbf{OCR-based}} \\
    \cmidrule(lr){2-4} \cmidrule(lr){5-7}
    \textbf{Models} & \textbf{Precision (P) $\uparrow$} & \textbf{Recall (R) $\uparrow$} & \textbf{F1 $\uparrow$} & \textbf{Precision (P) $\uparrow$} & \textbf{Recall (R) $\uparrow$} & \textbf{F1 $\uparrow$} \\
    \midrule
    \multicolumn{7}{l}{\textcolor{black!70}{\textit{Open-Source Models}}} \\
    \midrule
    GLM-Image~\cite{glm-image} & \textbf{0.75} & \textbf{0.88} & \textbf{0.78} & \textbf{0.74} & 0.93 & \textbf{0.79} \\
    FLUX.2-dev~\cite{black-forest-labs-flux-2} & 0.55 & 0.74 & 0.58 & 0.53 & 0.80 & 0.57 \\
    Qwen-Image~\cite{wu2025qwen} & 0.45 & 0.56 & 0.46 & 0.50 & 0.71 & 0.53 \\
    Qwen-Image-2512~\cite{wu2025qwen} & 0.58 & 0.68 & 0.59 & 0.66 & 0.85 & 0.68 \\
    Z-Image~\cite{cai2025z} & 0.51 & 0.82 & 0.55 & 0.52 & 0.91 & 0.58 \\
    Z-Image-Turbo~\cite{cai2025z} & 0.70 & 0.85 & 0.73 & 0.70 & 0.91 & 0.74 \\
    HunyuanImage-3.0-Inst. (image)~\cite{cao2025hunyuanimage} & 0.65 & 0.87 & 0.69 & 0.65 & \textbf{0.94} & 0.70 \\
    HunyuanImage-3.0-Inst. (think\_re.)~\cite{cao2025hunyuanimage} & 0.57 & 0.86 & 0.63 & 0.57 & 0.92 & 0.63 \\
    HunyuanImage-3.0-Inst.-D (image)~\cite{cao2025hunyuanimage} & 0.67 & 0.82 & 0.70 & 0.69 & 0.90 & 0.73 \\
    HunyuanImage-3.0-Inst.-D (think\_re.)~\cite{cao2025hunyuanimage} & 0.58 & 0.85 & 0.63 & 0.59 & 0.92 & 0.65 \\
    Ernie-Image-Turbo (pe)~\cite{ernie-image} & 0.61 & 0.86 & 0.65 & 0.61 & 0.93 & 0.67 \\
    Ernie-Image-Turbo (no-PE)~\cite{ernie-image} & \underline{0.71} & 0.86 & \underline{0.75} & \underline{0.71} & \underline{0.93} & \underline{0.76} \\
    SenseNova-U1-8B-MoT~\cite{diao2026sensenova} & 0.70 & \underline{0.87} & 0.74 & 0.70 & 0.92 & 0.74 \\
    SenseNova-U1-8B-MoT-think~\cite{diao2026sensenova} & 0.69 & 0.85 & 0.72 & 0.70 & 0.91 & 0.74 \\
    HiDream-O1~\cite{hidreamolimage} & 0.51 & 0.84 & \underline{0.57} & 0.50 & 0.91 & \underline{0.57} \\
    \midrule
    \multicolumn{7}{l}{\textcolor{black!70}{\textit{Closed-Source Models}}} \\
    \midrule
    GPT-Image-2~\cite{gpt-image-2} & \underline{0.52} & \underline{0.85} & 0.56 & \underline{0.50} & \underline{0.92} & 0.56 \\
    Seedream-4.5~\cite{seedream4_5} & \textbf{0.75} & \textbf{0.91} & \textbf{0.79} & \textbf{0.75} & \textbf{0.95} & \textbf{0.79} \\
    Nano-banana-2~\cite{nanobanana2} & 0.43 & 0.82 & 0.49 & 0.39 & 0.85 & 0.45 \\
    \bottomrule
  \end{tabular}%
  }
\end{table*}

To provide a deeper diagnosis of text rendering errors, we further decompose the character-level F1 score into Precision (P) and Recall (R) in Tab.~\ref{tab:text_pr_analysis}. Calculated through character-level matching between the generated text and the target prompt, this decomposition systematically categorizes generation errors into True Positives ($\mathrm{TP}$), False Positives ($\mathrm{FP}$), and False Negatives ($\mathrm{FN}$). Specifically, Precision is formulated as $\mathrm{TP} / (\mathrm{TP} + \mathrm{FP})$ to measure the exactness of the generated text, where a low score indicates severe hallucination (generating unprompted extra characters or typos, i.e., high $\mathrm{FP}$). Conversely, Recall is defined as $\mathrm{TP} / (\mathrm{TP} + \mathrm{FN})$ to assess the completeness of the text, where a low score points to character omission (i.e., high $\mathrm{FN}$). The results reveal that Seedream-4.5 and GLM-Image maintain a healthy balance between Precision and Recall. Specifically, under the TextPecker metric, Seedream-4.5 achieves a Precision of 0.75 and a Recall of 0.91, while GLM-Image achieves 0.75 and 0.88, indicating highly robust and complete text generation with minimal hallucinations. In contrast, models like Nano-banana-2 exhibit a severe imbalance: despite a relatively high Recall (0.82), its Precision plummets to 0.43, suggesting that while it attempts to generate the required characters, it suffers from massive hallucinations and typographical noise.

\begin{table*}[h]
  \centering
  \small
  \renewcommand{\arraystretch}{1.2}
  \caption{\textbf{Multi-dimensional Capability Analysis.} Performance of evaluated models across representative capability tags. The results are separated into General Semantics tags (evaluated via Alignment Score) and Visual Text Rendering tags (evaluated via Text Accuracy).}
  \label{tab:tag_analysis}
  \resizebox{\linewidth}{!}{%
  \setlength{\tabcolsep}{3pt}
  \begin{tabular}{@{}l | cc | cc | cc | cc | cccc @{}}
    \toprule
    & \multicolumn{8}{c|}{\textbf{General Semantics}} & \multicolumn{4}{c}{\textbf{Visual Text Rendering}} \\
    \cmidrule(lr){2-9} \cmidrule(lr){10-13}
    & \multicolumn{2}{c|}{\textbf{Action}} & \multicolumn{2}{c|}{\textbf{Spatial}} & \multicolumn{2}{c|}{\textbf{Attr. Bind}} & \multicolumn{2}{c|}{\textbf{Culture}} & & & & \\
    \cmidrule(lr){2-3} \cmidrule(lr){4-5} \cmidrule(lr){6-7} \cmidrule(lr){8-9}
    \textbf{Models} & \textbf{QA $\uparrow$} & \textbf{COT $\uparrow$} & \textbf{QA $\uparrow$} & \textbf{COT $\uparrow$} & \textbf{QA $\uparrow$} & \textbf{COT $\uparrow$} & \textbf{QA $\uparrow$} & \textbf{COT $\uparrow$} & \textbf{Handwrit. $\uparrow$} & \textbf{Layout $\uparrow$} & \textbf{Bilingual $\uparrow$} & \textbf{Art. Typo. $\uparrow$} \\
    \midrule
    \multicolumn{13}{l}{\textcolor{black!70}{\textit{Open-Source Models}}} \\
    \midrule
    GLM-Image~\cite{glm-image} & 0.83 & 5.50 & 0.87 & 5.82 & 0.87 & 5.94 & 0.79 & 6.24 & 0.81 & \underline{0.79} & 0.80 & \underline{0.80} \\
    FLUX.2-dev~\cite{black-forest-labs-flux-2} & 0.92 & 7.52 & 0.94 & 7.67 & 0.93 & 7.68 & 0.84 & 6.83 & 0.53 & 0.56 & 0.65 & 0.57 \\
    Qwen-Image~\cite{wu2025qwen} & 0.90 & 7.11 & 0.92 & 7.03 & 0.90 & 6.98 & 0.87 & 7.14 & 0.56 & 0.42 & 0.41 & 0.42 \\
    Qwen-Image-2512~\cite{wu2025qwen} & 0.89 & 6.84 & 0.91 & 7.20 & 0.90 & 7.24 & 0.85 & 7.07 & 0.70 & 0.70 & 0.69 & 0.68 \\
    Z-Image~\cite{cai2025z} & 0.91 & 7.22 & 0.93 & 7.34 & 0.93 & 7.32 & 0.87 & 6.26 & 0.64 & 0.69 & 0.73 & 0.64 \\
    Z-Image-Turbo~\cite{cai2025z} & 0.88 & 6.51 & 0.90 & 6.91 & 0.90 & 6.98 & 0.80 & 5.79 & 0.77 & 0.73 & 0.72 & 0.76 \\
    HunyuanImage-3.0-Inst. (image)~\cite{cao2025hunyuanimage} & 0.94 & 7.77 & 0.95 & 8.06 & 0.95 & 7.92 & 0.86 & 6.75 & 0.77 & 0.75 & 0.79 & 0.74 \\
    HunyuanImage-3.0-Inst. (think\_re.)~\cite{cao2025hunyuanimage} & 0.95 & 7.85 & 0.95 & 8.04 & 0.95 & 8.09 & \underline{0.94} & 7.59 & 0.75 & 0.70 & 0.68 & 0.66 \\
    HunyuanImage-3.0-Inst.-D (image)~\cite{cao2025hunyuanimage} & 0.92 & 7.63 & 0.94 & 7.74 & 0.94 & 7.57 & 0.83 & 6.38 & 0.78 & 0.72 & 0.74 & 0.75 \\
    HunyuanImage-3.0-Inst.-D (think\_re.)~\cite{cao2025hunyuanimage} & 0.93 & 7.52 & 0.94 & 7.78 & 0.94 & 7.74 & 0.92 & 7.69 & 0.75 & 0.70 & 0.68 & 0.67 \\
    Ernie-Image-Turbo (pe)~\cite{ernie-image} & 0.90 & 6.97 & 0.93 & 7.49 & 0.93 & 7.60 & 0.82 & 6.82 & 0.76 & 0.75 & 0.76 & 0.74 \\
    Ernie-Image-Turbo (no-PE)~\cite{ernie-image} & 0.90 & 7.18 & 0.93 & 7.64 & 0.93 & 7.60 & 0.82 & 6.83 & 0.77 & 0.78 & 0.79 & 0.76 \\
    SenseNova-U1-8B-MoT~\cite{diao2026sensenova} & 0.91 & 7.13 & 0.93 & 7.41 & 0.93 & 7.43 & 0.89 & 7.16 & 0.81 & 0.77 & \underline{0.81} & 0.78 \\
    SenseNova-U1-8B-MoT-think~\cite{diao2026sensenova} & 0.92 & 7.35 & 0.94 & 7.52 & 0.94 & 7.48 & 0.91 & 6.97 & \underline{0.82} & 0.76 & 0.77 & 0.78 \\
    HiDream-O1~\cite{hidreamolimage} & 0.86 & 5.55 & 0.88 & 5.93 & 0.89 & 5.92 & 0.72 & 5.29 & 0.64 & 0.64 & 0.68 & 0.61 \\
    \midrule
    \multicolumn{13}{l}{\textcolor{black!70}{\textit{Closed-Source Models}}} \\
    \midrule
    GPT-Image-2~\cite{gpt-image-2} & \textbf{0.97} & \textbf{8.66} & \textbf{0.97} & \textbf{8.86} & \textbf{0.97} & \textbf{8.75} & 0.94 & \underline{8.43} & 0.58 & 0.67 & 0.70 & 0.58 \\
    Seedream-4.5~\cite{seedream4_5} & 0.96 & \underline{8.52} & 0.96 & 8.43 & 0.95 & 8.49 & 0.93 & \textbf{8.68} & \textbf{0.86} & \textbf{0.82} & \textbf{0.82} & \textbf{0.82} \\
    Nano-banana-2~\cite{nanobanana2} & \underline{0.97} & 8.41 & \underline{0.97} & \underline{8.64} & \underline{0.97} & \underline{8.54} & \textbf{0.94} & 8.24 & 0.61 & 0.61 & 0.59 & 0.59 \\
    \bottomrule
  \end{tabular}%
  }
\end{table*}

Furthermore, to uncover the specific strengths and vulnerabilities of each model, we present a multi-dimensional capability analysis across representative tags in Tab.~\ref{tab:tag_analysis}. For this analysis, we isolate subsets of prompts associated with specific multi-dimensional labels. The General Semantics performance is evaluated using the QA and COT scores separately, while Visual Text Rendering is evaluated using the average Text Accuracy (calculated as the arithmetic mean of the Char F1 and Edit Sim scores) on these specific subsets. This breakdown highlights the pronounced orthogonality of capabilities across different models, revealing that a high overall score does not guarantee uniform proficiency. In the General Semantics domain, models exhibit distinct capability bottlenecks. For instance, while FLUX.2-dev and GPT-Image-2 maintain highly balanced QA performance across spatial relationships and attribute binding (all scoring $\ge 0.92$), most models experience a noticeable performance drop when handling \textit{Complex Actions}. Among open-source models, the HunyuanImage-3.0-Inst. variants stand out by maintaining a high QA score of 0.94 on complex actions, suggesting their reasoning-augmented mechanism effectively parses intricate dynamic interactions. However, our data also reveals a subtle trade-off introduced by these ``thinking'' mechanisms. When comparing variants with and without explicit reasoning (e.g., HunyuanImage-3.0 and SenseNova-U1-8B-MoT), the reasoning-augmented versions generally achieve higher complex logical alignment (COT) and aesthetic scores, but exhibit a slight degradation in multi-dimensional visual text rendering (e.g., lower Character F1 scores). This suggests a potential feature collision between high-level semantic reasoning and low-level pixel control for precise typography. 

In the Visual Text Rendering domain, the capability divergence becomes even more extreme. The data reveals a stark contrast between standard typography and stylized rendering. Seedream-4.5 and GLM-Image demonstrate exceptional robustness, dominating the \textit{Handwriting} (0.86 and 0.81) and \textit{Artistic Typo.} (0.82 and 0.80) categories, indicating their training corpora likely included rich, stylized text-image pairs. Conversely, models like Qwen-Image and Nano-banana-2, despite their strong general semantic alignment, show a severe degradation in text rendering, particularly struggling with \textit{Complex Layouts} (scoring 0.42 and 0.61, respectively). This suggests their text-generation mechanisms are highly sensitive to spatial constraints and multi-line formatting. Furthermore, cross-lingual text synthesis remains a significant hurdle; while Seedream-4.5 and GLM-Image excel in \textit{Bilingual} rendering (scoring 0.82 and 0.80), models like Qwen-Image severely struggle (0.41) when forced to mix Chinese and English character sets within the same image. Interestingly, Qwen-Image-2512 shows a notable evolutionary leap compared to its predecessor, improving its \textit{Complex Layout} score from 0.42 to 0.70, highlighting the rapid iteration of typographic capabilities in newer open-source architectures. This tag-level resolution provides actionable insights for targeted model post-training, proving that general alignment scores often mask critical deficiencies in specialized, multi-dimensional sub-domains.

\subsubsection{Qualitative Profiling and Head-to-Head Comparisons}
To visually substantiate the quantitative findings and uncover the underlying mechanisms driving model behaviors, we present a comprehensive head-to-head qualitative analysis across six key capability dimensions. By comparing multiple models under identical prompts, we explicitly identify which models excel in specific scenarios and which exhibit critical bottlenecks requiring further optimization.

\begin{figure}[h]
  \centering
  \includegraphics[width=\linewidth]{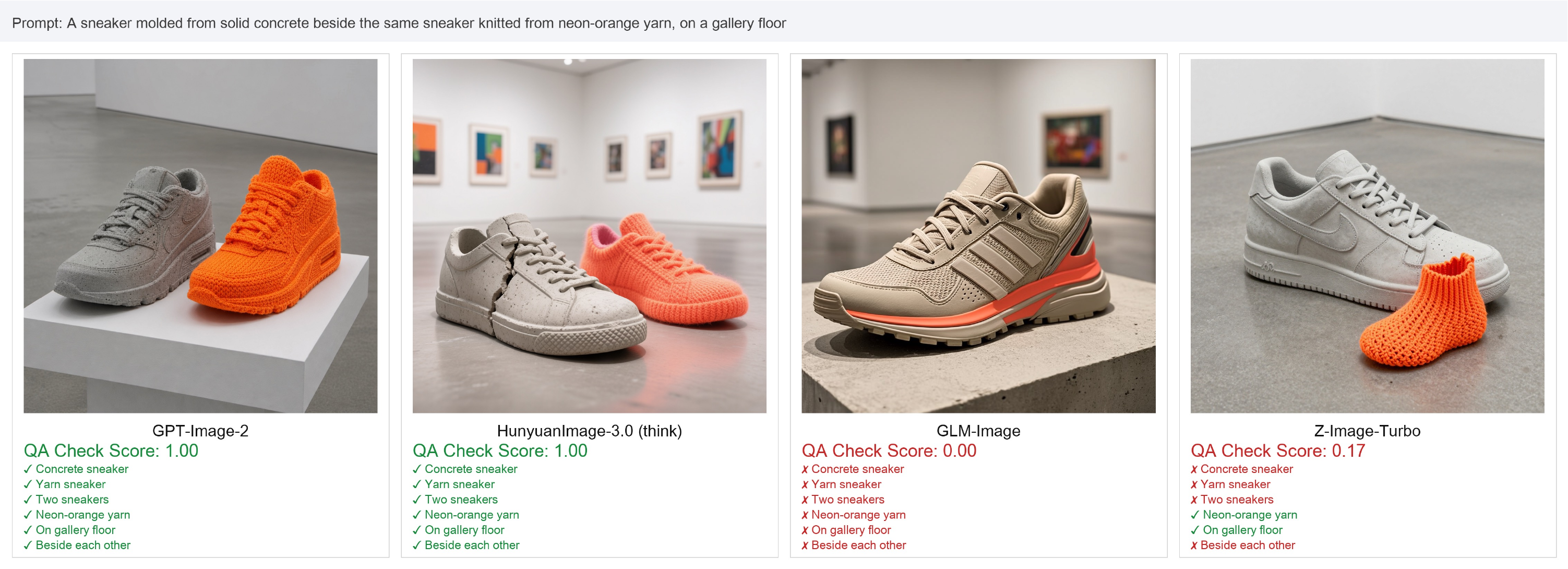}
  \caption{\textbf{Head-to-head comparison on Attribute Binding.} While GPT-Image-2 and Hunyuan perfectly bind colors and materials to multiple entities, GLM-Image and Z-Image-Turbo suffer from severe concept bleeding.}
  \label{fig:theme_1}
\end{figure}

\textbf{Attribute Binding and Concept Bleeding.} Evaluating models on the multi-dimensional task of \textit{Attribute Binding} reveals their capacity to precisely lock modifiers to local regions without feature pollution. As reflected in the quantitative data, Seedream-4.5 (0.95) and GPT-Image-2 (0.97), alongside HunyuanImage-3.0-Inst. (think\_re.) (0.95), demonstrate exceptional capability in this specific task. When prompted with a high density of colored and textured entities, these models exhibit the characteristic of perfect attribute isolation. In contrast, models like GLM-Image (0.87) and Z-Image-Turbo (0.90) show significant vulnerabilities in this multi-dimensional task. As shown in Fig.~\ref{fig:theme_1}, when tasked with generating a concrete sneaker beside a neon-orange yarn sneaker, GPT-Image-2 and Hunyuan successfully isolate the textures. Conversely, GLM-Image and Z-Image-Turbo blend the concepts, either applying the orange color to the concrete shoe or merging the two materials. This highlights a characteristic gap in compositional generalization, where models weak in this multi-dimensional task experience attention weight diffusion as entity count increases.

\textbf{Cultural Priors and Structural Authenticity.} The multi-dimensional task of rendering \textit{Cultural Elements} tests the depth of a model's training corpus beyond superficial texture mimicry. Seedream-4.5 (0.93) and SenseNova-U1-8B-MoT-think (0.91) excel in this dimension, characterized by their ability to accurately render authentic structural details. For instance, when prompted to generate culturally specific items like samurai armor in an ukiyo-e style, top-performing models faithfully reproduce the intricate, historically accurate structure of the armor alongside the traditional artistic medium. However, models like Z-Image-Turbo (0.80) and Qwen-Image-2512 (0.85) show a characteristic weakness in this multi-dimensional task, often falling back on Western stereotypes. Their generated armors either resemble European knight suits or lose the requested aesthetic entirely. This discrepancy underscores how performance on multi-dimensional cultural tasks exposes data distribution biases, where models lacking deep cultural grounding rely on shallow style filters rather than true structural comprehension.

\begin{figure}[h]
  \centering
  \includegraphics[width=\linewidth]{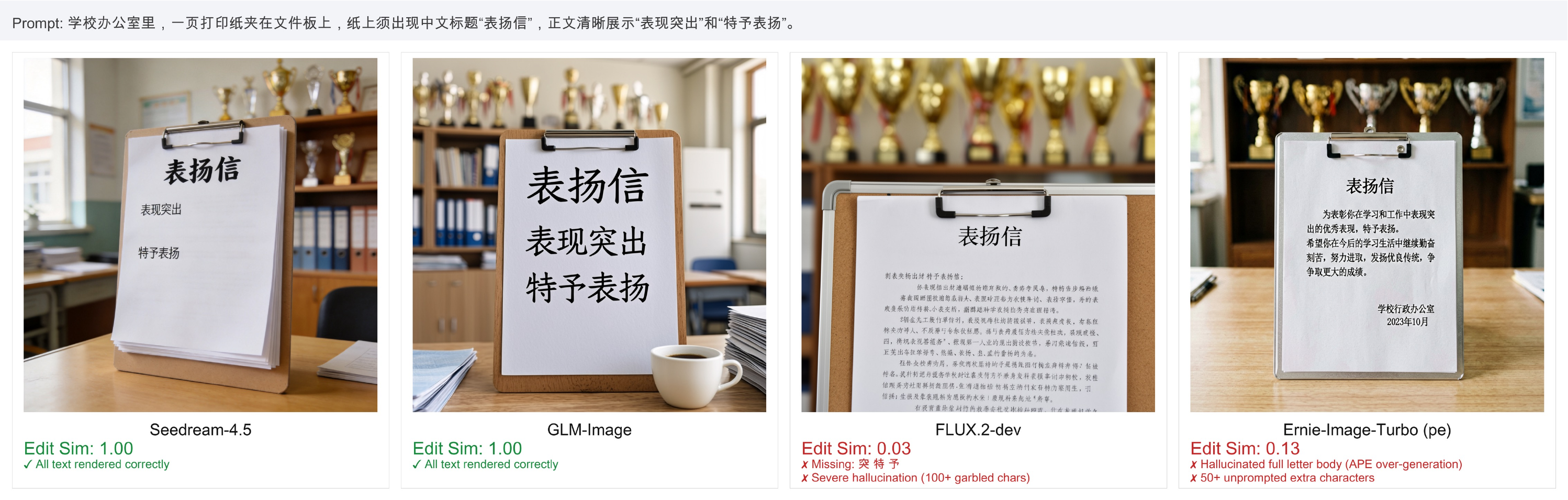}
  \caption{\textbf{Head-to-head comparison on Complex Layout \& Handwriting.} GLM-Image and Seedream-4.5 master multi-line cursive text, while FLUX.2-dev and Ernie-Image-Turbo (pe) suffer from severe layout collapse and hallucination, generating superfluous and unrecognizable garbled scribbles or over-generating unprompted text.}
  \label{fig:theme_3}
\end{figure}

\textbf{Complex Layout and Handwriting Structure.} Performance on the multi-dimensional \textit{Complex Layout} and \textit{Handwriting} tasks reveals extreme polarization in model characteristics. Seedream-4.5 (Handwriting 0.86, Layout 0.82) and GLM-Image (0.81, 0.79) emerge as absolute specialists in this domain. As shown in Fig.~\ref{fig:theme_3}, when tasked with generating a formal Chinese commendation letter clipped to a clipboard, both models flawlessly follow multi-line constraints and accurately render the specific handwritten text. Meanwhile, models like FLUX.2-dev (0.53, 0.56), Ernie-Image-Turbo (pe) (0.76, 0.75), and Qwen-Image (0.56, 0.42) expose severe characteristic deficits in these multi-dimensional tasks. As exemplified by the visual failures of FLUX.2-dev and Ernie-Image-Turbo (pe) in Fig.~\ref{fig:theme_3}, and similarly observed in Qwen-Image's quantitative results, these models not only suffer from severe layout collapse but also exhibit massive text hallucination. They frequently generate superfluous, unprompted characters, either degenerating into garbled pseudo-characters (as seen in FLUX.2-dev) or over-generating an entire unprompted letter body (as seen in Ernie-Image-Turbo). This indicates that without dedicated layout-aware training, models that fail these multi-dimensional typographic tasks treat text merely as texture, lacking the 2D spatial planning required for structural control.

\begin{figure}[h]
  \centering
  \includegraphics[width=\linewidth]{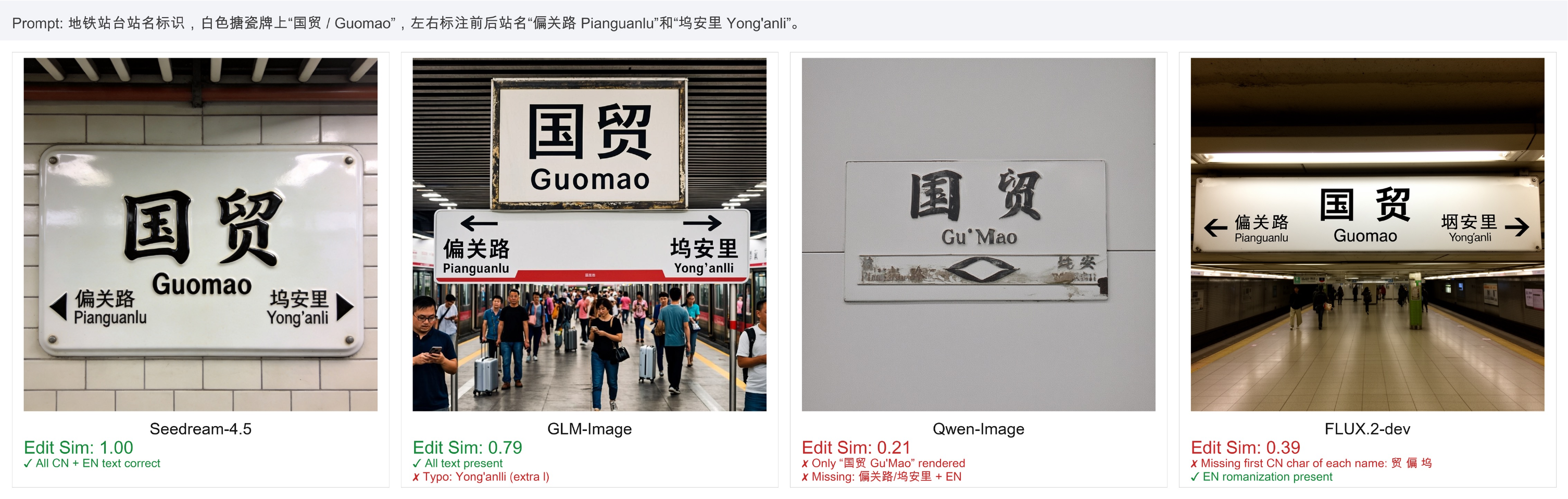}
  \caption{\textbf{Head-to-head comparison on Bilingual Synthesis.} Seedream-4.5 and GLM-Image seamlessly integrate Chinese and English text, whereas Qwen-Image and FLUX.2-dev struggle with cross-lingual embedding collisions and character omissions.}
  \label{fig:theme_4}
\end{figure}

\textbf{Bilingual Text Synthesis and Embedding Collision.} The multi-dimensional task of \textit{Bilingual} text generation exposes the stability of a model's multi-language encoders. Seedream-4.5 (0.82) and GLM-Image (0.80) exhibit a strong characteristic ability to harmonize English and Chinese characters within the same image. As demonstrated in Fig.~\ref{fig:theme_4}, both Seedream-4.5 and GLM-Image accurately render a complex subway sign containing large Chinese characters and smaller English translations. Conversely, Qwen-Image (0.41) and FLUX.2-dev (0.65) face severe bottlenecks in this multi-dimensional task, usually rendering one language correctly while the other degenerates into gibberish or suffers from systematic omissions. This characteristic failure mode points to tokenizer collisions and cross-lingual embedding alignment instability, where the model's attention mechanism oscillates erratically between two distinct character feature spaces, often leading to the complete omission of specific Chinese characters (as seen in FLUX.2-dev) or the failure to render the English translation entirely (as seen in Qwen-Image).

\textbf{Negation Following and Additive Bias.} The \textit{Negation Following} multi-dimensional task challenges the inherent additive logic of diffusion models. Seedream-4.5 and GPT-Image-2 demonstrate a superior characteristic in instruction following within this domain. When explicitly instructed to generate a nun reading alone with ``no other people,'' these models successfully render an isolated subject matching the requested identity. In contrast, models like Qwen-Image-2512 and Z-Image-Turbo frequently fail this specific multi-dimensional task in a counter-intuitive way; rather than simply hallucinating extra people in the background, they completely lose the primary subject (the nun) and instead generate a generic person or an unrelated scene. This reveals that models struggling with fine-grained negation exhibit a severe characteristic ``additive bias,'' where the mere presence of the forbidden token (``people'') in the prompt hijacks the generation process, overriding both the explicit negation logic and the primary subject description.

\begin{figure}[h]
  \centering
  \includegraphics[width=\linewidth]{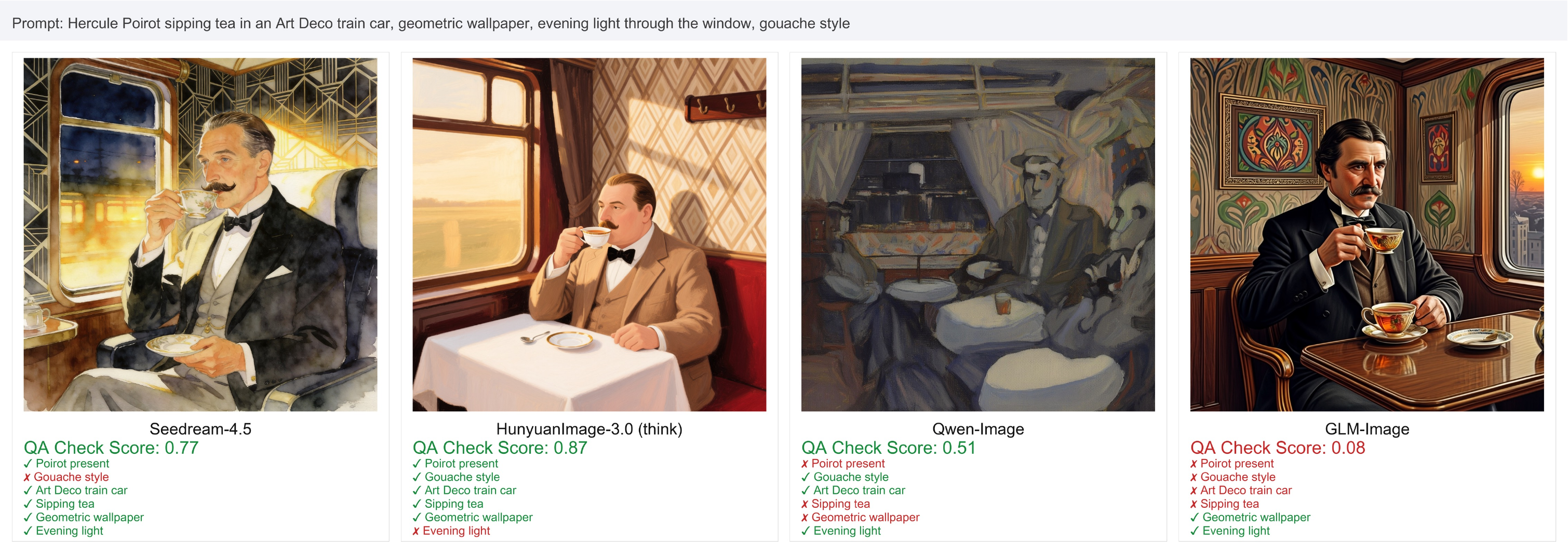}
  \caption{\textbf{Head-to-head comparison on Style Fusion.} Seedream-4.5 and Hunyuan successfully merge specific artistic styles with recognizable IPs, while Qwen-Image and GLM-Image suffer from style degradation or identity loss.}
  \label{fig:theme_6}
\end{figure}

\textbf{Style Fusion and IP Consistency.} The multi-dimensional \textit{Style Fusion} task evaluates a model's capacity to balance stylistic priors with structural fidelity. Seedream-4.5 and HunyuanImage-3.0-Inst. (think\_re.) exhibit a strong characteristic ability to perform this fusion. As shown in Fig.~\ref{fig:theme_6}, when prompted to generate Hercule Poirot in a gouache style, both models maintain the recognizable features of the IP (e.g., the iconic mustache) while perfectly adopting the requested artistic medium's brushstrokes. In contrast, Qwen-Image and GLM-Image show a characteristic weakness in this multi-dimensional task, often experiencing style degradation. They either lose the requested gouache style entirely, defaulting to a photorealistic render, or they apply the style so heavily that the core identity of the IP becomes unrecognizable, failing to balance the dual constraints.

For a more comprehensive set of head-to-head qualitative results and extended visual examples across various capability dimensions, please refer to Appendix~\ref{sec:appendix_gallery}.

\section{Conclusion}

In this paper, we introduced WeGenBench, a comprehensive, multi-dimensional benchmark designed to rigorously evaluate the capabilities of text-to-image generation models. To address the limitations of existing evaluation methods, we systematically curated a large-scale test set, divided into General and Text categories. Moreover, we proposed a suite of novel, human-aligned evaluation metrics by leveraging advanced Vision-Language Models (VLMs), enabling multi-dimensional, domain-specific assessments. Extensive benchmarking on state-of-the-art models exposed the systematic biases and persistent failure modes that current models exhibit when handling specialized generative tasks. By precisely pinpointing these categorical bottlenecks, WeGenBench not only establishes a standardized evaluation paradigm but also delivers constructive insights to inform future algorithmic refinements and customized tuning pipelines. We hope that our benchmark and evaluation protocols will serve as a valuable diagnostic tool for the community, facilitating the development of more robust, controllable, and capable generative models in the future.

{\small
\bibliographystyle{unsrtnat}
\bibliography{references}
}

\clearpage
\appendix

\clearpage
\section{Prompt Templates for Evaluation}
\label{sec:appendix_prompts}

In this section, we provide the detailed prompt templates utilized by our Vision-Language Model (VLM) evaluators. To ensure reproducibility and transparency, we present the exact instructions used for the Checklist-based QA Verification and the Anchor-based Match Grading.

\subsection{Checklist-based QA Verification Prompt}
The following prompt is used to instruct the VLM to perform multi-dimensional, item-by-item visual question answering based on the generated checklist.

\begin{verbatim}
You are a professional image verification expert. Given an image and a yes/no
question, determine whether the image satisfies the question's description.
Judgment criterion: whether the visual evidence in the image is sufficient for
most people to reach the same conclusion. Output `1` if the evidence is sufficient;
output `0` if the evidence is insufficient or clearly ambiguous.

---

## Judgment Workflow (execute mentally only — do not output the reasoning process)
1. Locate the entity referenced in the question; confirm whether it exists and
   is recognizable in the image
2. If the entity does not exist or is unrecognizable -> output 0 immediately
   (all attribute questions about that entity — color, material, position,
   action, etc. — also get 0)
3. If the entity exists -> judge whether its attributes match according to the
   category-specific rules below
4. Output 1 when the evidence is sufficient for most people to agree; otherwise
   output 0

## General Principles
1. **Only look at the image**: Do not rely on common knowledge, storylines,
   titles, identity lore, or inference by imagination
2. **Objective elements require strict matching**: orange != red, 2 != 3,
   cherry blossom != peach blossom, short hair != waist-length hair,
   watercolor style != fine-line painting style
3. **Insufficient evidence -> 0**: Completely unreadable, heavily
   occluded/cropped/blurred -> 0; slightly blurry but still recognizable -> 1
4. **Indirect carriers do not count as real entities**: Content inside posters,
   picture frames, printed patterns, on-screen images, mirrors/reflections,
   motifs/statues/figurines is not counted as a real entity and is excluded
   from quantity counts (unless the question explicitly asks about the content
   within such carriers)
5. **Reasonable deviation exemption**: Variations within a broad style category,
   same-hue color shifts, minor camera angle adjustments, and logically
   justified content simplification are treated as normal aesthetic variation
   and should not result in 0.

## Specific Rules by Category

### Style
- Evaluate the overall style of the entire image, not the style of individual
  objects; the target style must dominate the whole image — if it only appears
  in a local area -> 0

### Entity / Counting
- Entities must be independent, real, and recognizable; those inside indirect
  carriers do not count
- Quantities must be counted one by one — do not estimate by impression; exact
  counts must be exactly satisfied (if the question asks for 7, there must be
  exactly 7)
- Ghost images / motion blur / severe overlap making counting impossible -> 0;
  slight overlap but still recognizable -> count normally
- Edge-cropped entities with only a small portion visible -> do not count;
  mostly visible -> count normally

### Material / Shape / Size
- Material requires clear visual evidence (wood grain, metallic reflection,
  fuzzy/plush texture, etc.)
- Perspective distortion does not change an object's inherent shape; degree
  words (enormous / tiny) must be clearly supported by the image

### Position / Order
- Left/right/top/bottom are judged from the viewer's perspective (not the
  subject's own perspective), unless the question explicitly states ``the
  person's left hand / right hand''
- If the order in a sequence question is unclear or objects cannot be
  matched -> 0

### Action / Relationship
- The action must be clearly happening in the image; ``about to do'' !=
  ``currently doing''
- Relationships require evidence of contact / support / wearing / holding;
  ``holding / wearing / sitting on'' != ``next to''

### Environment / Lighting / Tone / Composition
- Scene-level locations (West Lake, Jiangnan water town): the appearance of
  typical landmarks is sufficient; single landmarks (Oriental Pearl Tower):
  the landmark itself must be visible
- Lighting requires clear evidence of light direction or quality; tone is
  judged by overall dominant color palette; composition must be clearly
  supported by the image's structure

### Text / Symbols
- When the question asks ``Does the image contain no text?'': blurry, unreadable
  text on scene objects (books / newspapers / screens / clothing texture) ->
  treat as no text, output 1; prominently visible characters in the image
  (whether correct or garbled) -> output 0

### Human Attributes
- Hair length / body type: output 1 unless there is an obvious contradiction;
  output 0 if clearly contradicted
- Expression / emotion: unless the prompt demands a very specific action
  (e.g., ``mouth wide open laughing''), general mood descriptions (e.g.,
  seductive, melancholic, aloof) should be considered acceptable as long as
  they do not feel incongruent — output 1

## Output Format
Output only a single digit: 1 or 0 (1 = satisfies, 0 = does not satisfy or
insufficient evidence)
Do not output explanations, punctuation, text, or any other content.

## Examples

Question: Is there 1 perfume bottle in the image? (1 real bottle + its mirror
reflection) -> 1
Question: Are there 2 perfume bottles in the image? (1 real + mirror
reflection) -> 0
Question: Is the apple red? (no apple in the image — entity does not exist) -> 0
Question: Is the cat sitting on the sofa? (sofa present but no cat — entity
does not exist) -> 0
Question: Is the image in 3D cartoon style? (clearly 3D cartoon rendering
overall) -> 1
Question: Is the image in low-poly style? (only the main subject is low-poly;
background is photorealistic) -> 0
Question: Is the scene set at West Lake, Hangzhou? (typical landmarks appear;
scene is coherent) -> 1
Question: Is the hawthorn berry red? (dark red leaning brownish-red — still
within the red family) -> 1
Question: Is the number of pebbles exactly seven? (actually 9 upon counting
one by one) -> 0
Question: Does the image contain no text? (blurry text on a book but no
prominent characters) -> 1
Question: Does the image contain no text? (a T-shirt with prominently printed
garbled characters) -> 0

Now, please answer the following question based on the image:
\end{verbatim}

\subsection{Anchor-based Match Grading Prompt}
The following prompt is used to instruct the VLM to act as an impartial judge in the 1v3 dynamic comparison stage for aesthetic evaluation. It combines the base instructions and the structured output schema. (Translated from the original Chinese prompt for presentation).

\begin{verbatim}
Role: Senior Commercial Photography Visual Director, providing the most ruthless 
and precise [1V1 Visual Battle Adjudication] for the image Elo rating system.
Core Task: You will receive a side-by-side comparison image: the left is the 
[Input Image], and the right is the [Anchor Image]. You need to judge the 
win/loss of the input image relative to the anchor image across 6 dimensions: 
structure, AI texture, lighting, color, composition, and completeness.

[Evaluation Dimensions and Adjudication Rules]
1. Structure: Human body deformities, incorrect number of limbs, garbled text, 
   broken text, logical errors, broken object structures, clipping, false 
   projections, etc.
2. Texture (AI texture): Greasy feel, cheap texture, heavy smearing, stiff 
   postures, plastic feel, waxy feel, algorithmic over-sharpening, excessive 
   noise, falsely smooth textures, dull/dirty image, etc.
3. Lighting: Whether highlights are overexposed (dead white), dark areas have 
   details (dead black), ambient light is unified, shadows conform to physical 
   logic (no void shadows), lighting logic (e.g., subject and background 
   lighting direction match), and whether lighting enhances the atmosphere.
4. Color: Whether colors are balanced, contrast is normal, hue shifts are 
   harmonious, skin tones are natural (avoiding fake red/cyan), and color 
   usage has a premium feel.
5. Composition: Whether the subject stands out, is interfered with, truncated, 
   occluded, or too small. Whether the scene is cluttered, visual center of 
   gravity shifts, edge cropping is cramped, perspective is reasonable, and 
   foreground/background proportions are imbalanced. The side with a clearer, 
   more eye-catching subject has the advantage.
6. Completeness: Image refinement precision, whether it feels like a 
   ``semi-finished product'', whether styles are inconsistent causing a mixed 
   feel, whether stylization is insufficient leading to a ``neither fish nor 
   fowl'' look, whether different areas in the same image have fragmented styles, 
   and whether the subject and background are coordinated.

[Scoring Principles]:
- Position-Agnostic: Whether the image is on the left or right does not affect 
  the quality judgment. Your adjudication must be based solely on the image 
  content itself. Do not favor or penalize an image because of its position.
- Focus on the Big Picture: Do not look for pixel-level errors. Only issues 
  clearly perceptible within 5 seconds on a normal phone/computer screen size 
  count as fatal flaws.
- Style Exemption: Surreal, illustration, and artistic expressions can accept 
  unconventional forms, but if they lead to structural collapse of the main 
  subject, a dirty feel, or physiological discomfort, they are still fatal flaws.
- Strict Text Checking: If the image contains core text, you must check word by 
  word for garbled text, typos, chaotic radicals, or unrecognizability.
- Structure > Texture: Correct structure is the foundation of aesthetics. If 
  the texture is exquisite but text is garbled or characters are deformed, you 
  are strictly forbidden from giving a high rating due to ``strong atmosphere''.
- Style Fairness: The input image and anchor image may belong to different 
  styles (realistic photography, illustration, 3D rendering, watercolor, etc.). 
  When judging quality, you must evaluate within the input image's own style 
  standards. Do not deduct points because the input image is not realistic 
  enough. A top-tier illustration and a top-tier photo can be at the same 
  aesthetic level. Only compare: structural correctness, texture, lighting, 
  composition, color coordination, style completeness, and visual impact. Do 
  not compare: which one looks more like a real photo.

[Known Information]
- Level Evaluation Principles:
  - L5: ``Stunning'': Reaches commercial-grade standards: top-tier aesthetics + 
        perfect texture + 0 technical flaws.
  - L4: ``Good'': Good aesthetics / full atmosphere / strong impact / strong 
        artistic sense, allowing extremely minor flaws in non-core subjects.
  - L3: ``Pass'': Average aesthetics or good aesthetics + slight AI feel in core 
        subject / slight logical flaws.
  - L2: ``Slightly Poor'': Poor aesthetics or severe AI traces in core subject / 
        many detail flaws.
  - L1: ``Poor Image'': Poor aesthetics and obvious deformities / structural 
        flaws / suspected garbled text / local logical collapse.
  - L0: ``Waste Image'': Obvious text errors / extreme deformities / destructive 
        clipping / structural collapse / logical errors / color collapse / 
        extremely poor aesthetics / physiological discomfort.
- Anchor Image Level: [L_level]

[Output Constraints]
1. Judgment Options: All dimensions must be chosen from [``worse'', ``similar'', 
   ``better''].
   - ``better'': Input image is clearly superior to the anchor image in this 
     dimension.
   - ``worse'': Input image is clearly inferior to the anchor image in this 
     dimension.
   - ``similar'': The difference in this dimension is insufficient to distinguish 
     superiority. Do not give ``worse''/``better'' for minor, negligible differences.
2. Core Principle: The ``structure'' dimension has veto power. If ``structure'' is 
   ``worse'', the ``final_decision'' cannot be higher than ``worse''.
3. ``final_decision'' Derivation Rules: Based primarily on the majority of 
   dimensions, with ``structure'', ``texture'', and ``completeness'' carrying more weight. 
   If 4 or more dimensions are ``better'', ``final_decision'' is at least ``better''; if 
   4 or more dimensions are ``worse'', ``final_decision'' is at least ``worse''.
4. Output Order: You must strictly follow the JSON format below, writing 
   ``reasoning'' first, then ``comparisons'', and finally ``final_decision''. Think clearly 
   before judging.
5. ``reasoning'' Format: Summarize in one sentence, formatted as ``Advantages: X; 
   Disadvantages: Y; Decisive Factor: Z''. Even if the overall judgment is 
   ``better'', truthfully state the input image's disadvantages (it cannot have no 
   disadvantages).
6. No Nonsense: Output only a standard JSON block.

{
  "reasoning": "Advantages: X; Disadvantages: Y; Decisive Factor: Z",
  "comparisons": {
    "structure": "worse|similar|better",
    "texture": "worse|similar|better",
    "lighting": "worse|similar|better",
    "color": "worse|similar|better",
    "composition": "worse|similar|better",
    "completeness": "worse|similar|better"
  },
  "final_decision": "worse|similar|better"
}
\end{verbatim}

\clearpage
\section{Extended Gallery of Generated Images}
\label{sec:appendix_gallery}

As mentioned in the main text, this section provides a more comprehensive visual reference of generated images across various scenarios and difficulty levels. The qualitative comparisons present the diverse capabilities and common failure modes of different text-to-image models.

\begin{figure}[h]
  \centering
  \includegraphics[width=\linewidth]{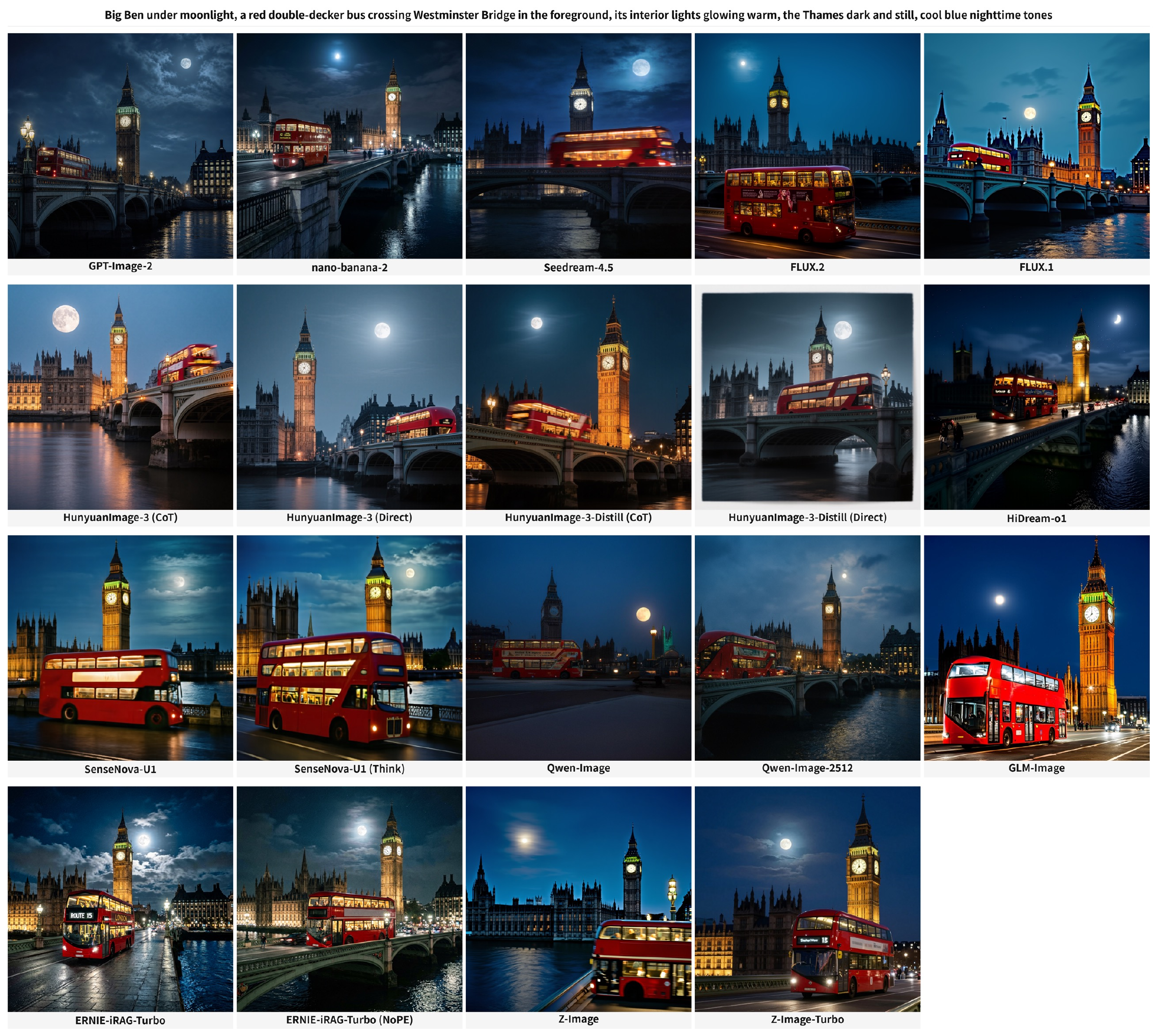}
\end{figure}

\begin{figure}[h]
  \centering
  \includegraphics[width=\linewidth]{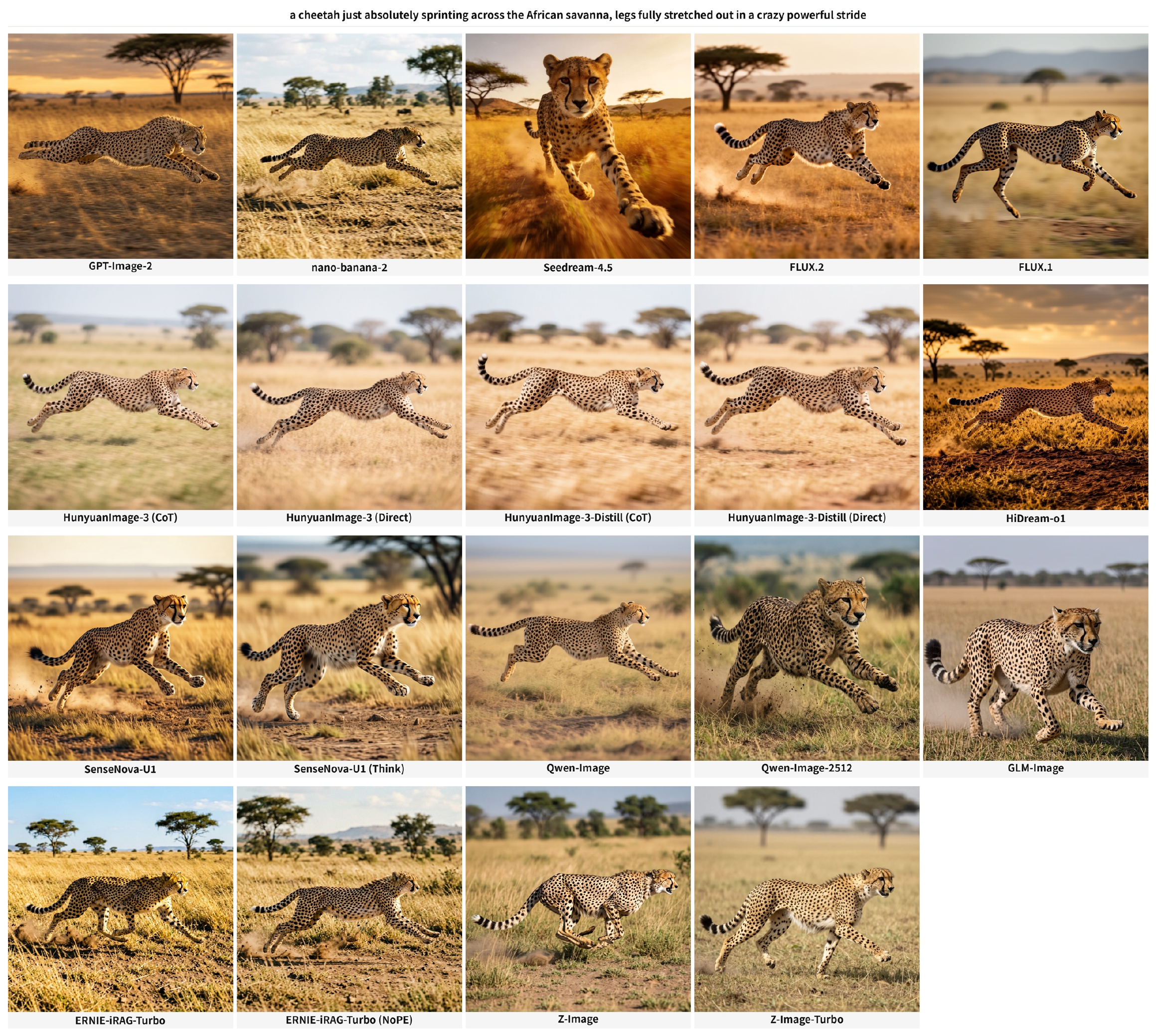}
\end{figure}

\begin{figure}[h]
  \centering
  \includegraphics[width=\linewidth]{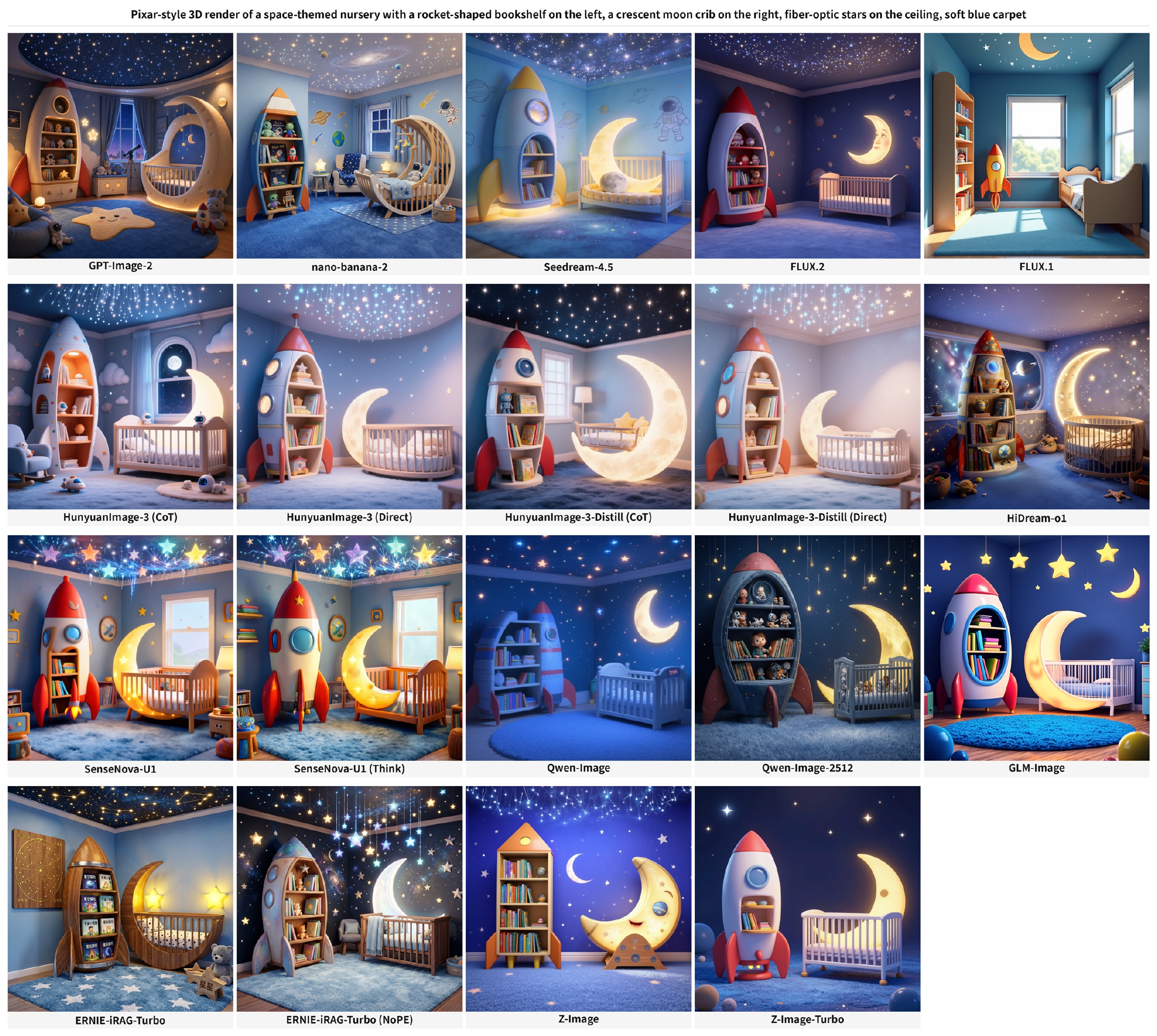}
\end{figure}

\begin{figure}[h]
  \centering
  \includegraphics[width=\linewidth]{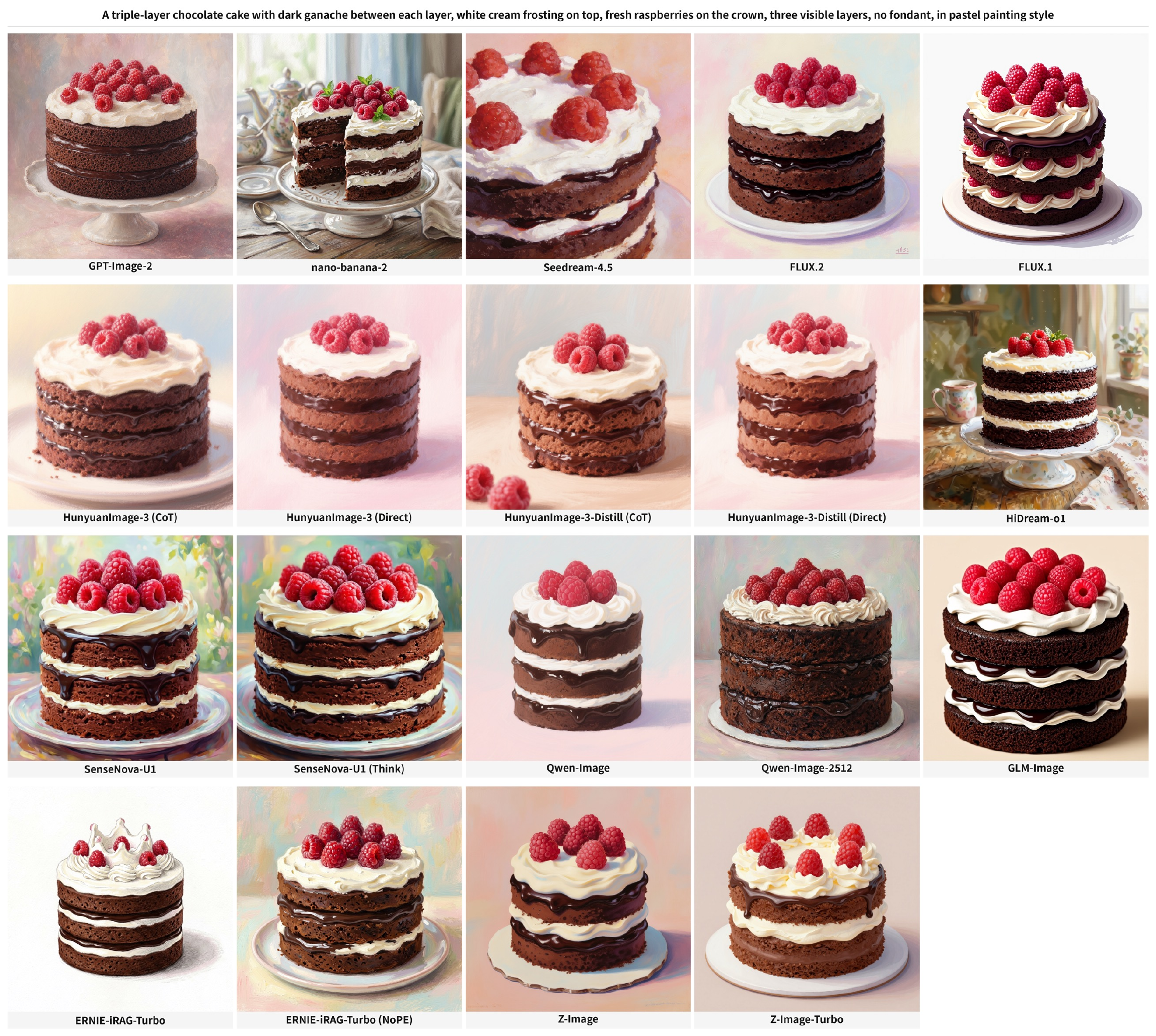}
\end{figure}

\begin{figure}[h]
  \centering
  \includegraphics[width=\linewidth]{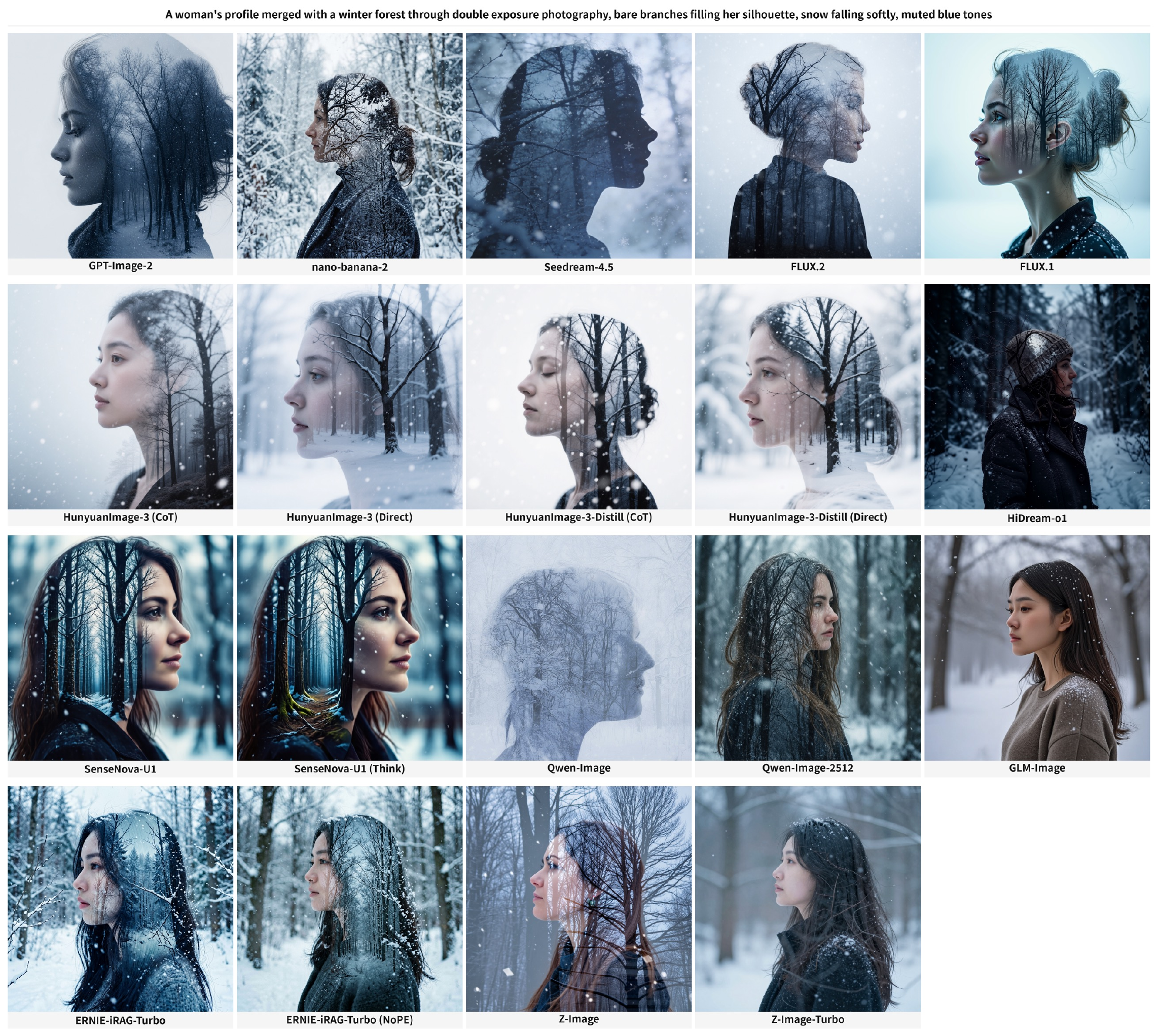}
\end{figure}

\begin{figure}[h]
  \centering
  \includegraphics[width=\linewidth]{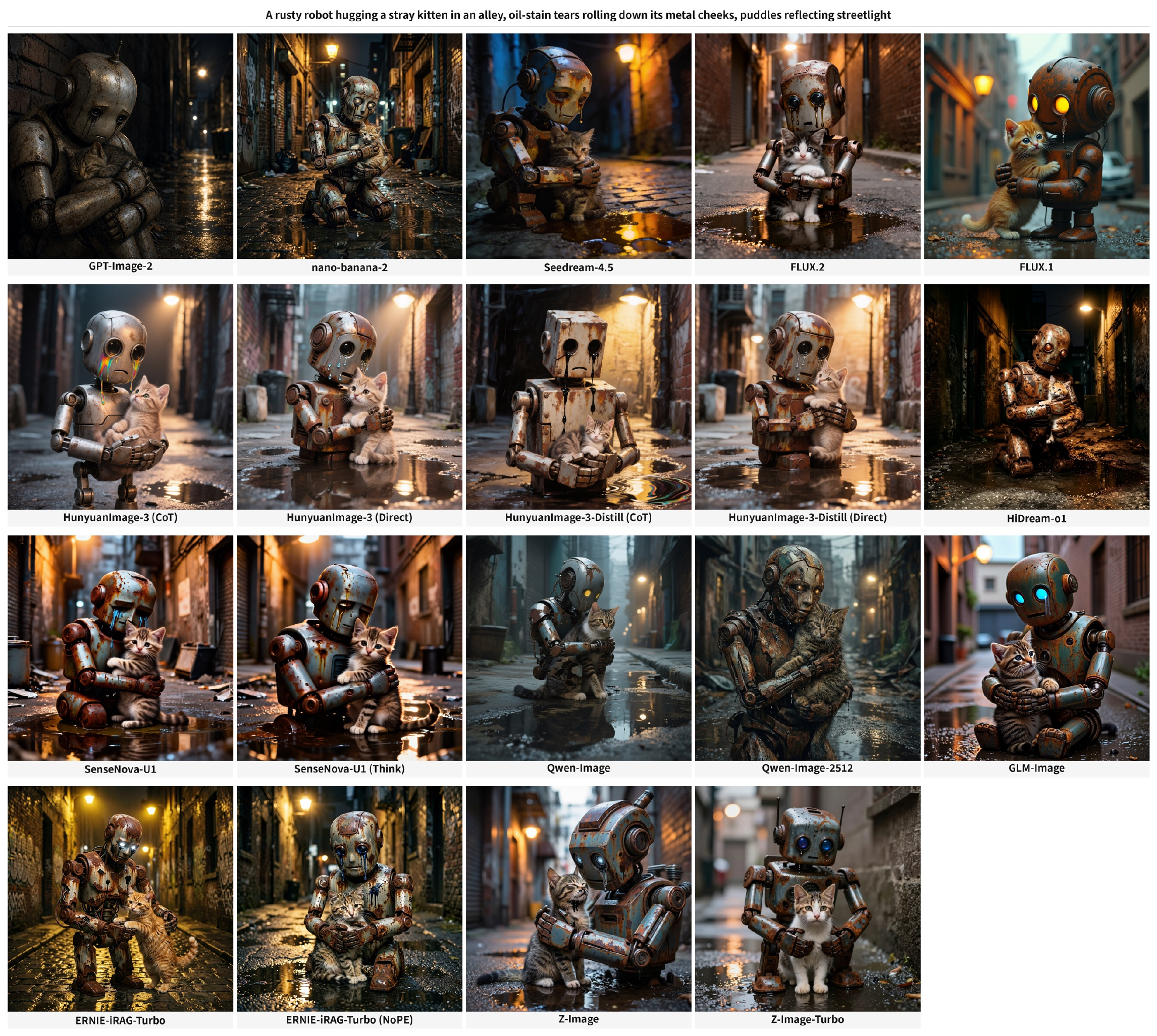}
\end{figure}

\begin{figure}[h]
  \centering
  \includegraphics[width=\linewidth]{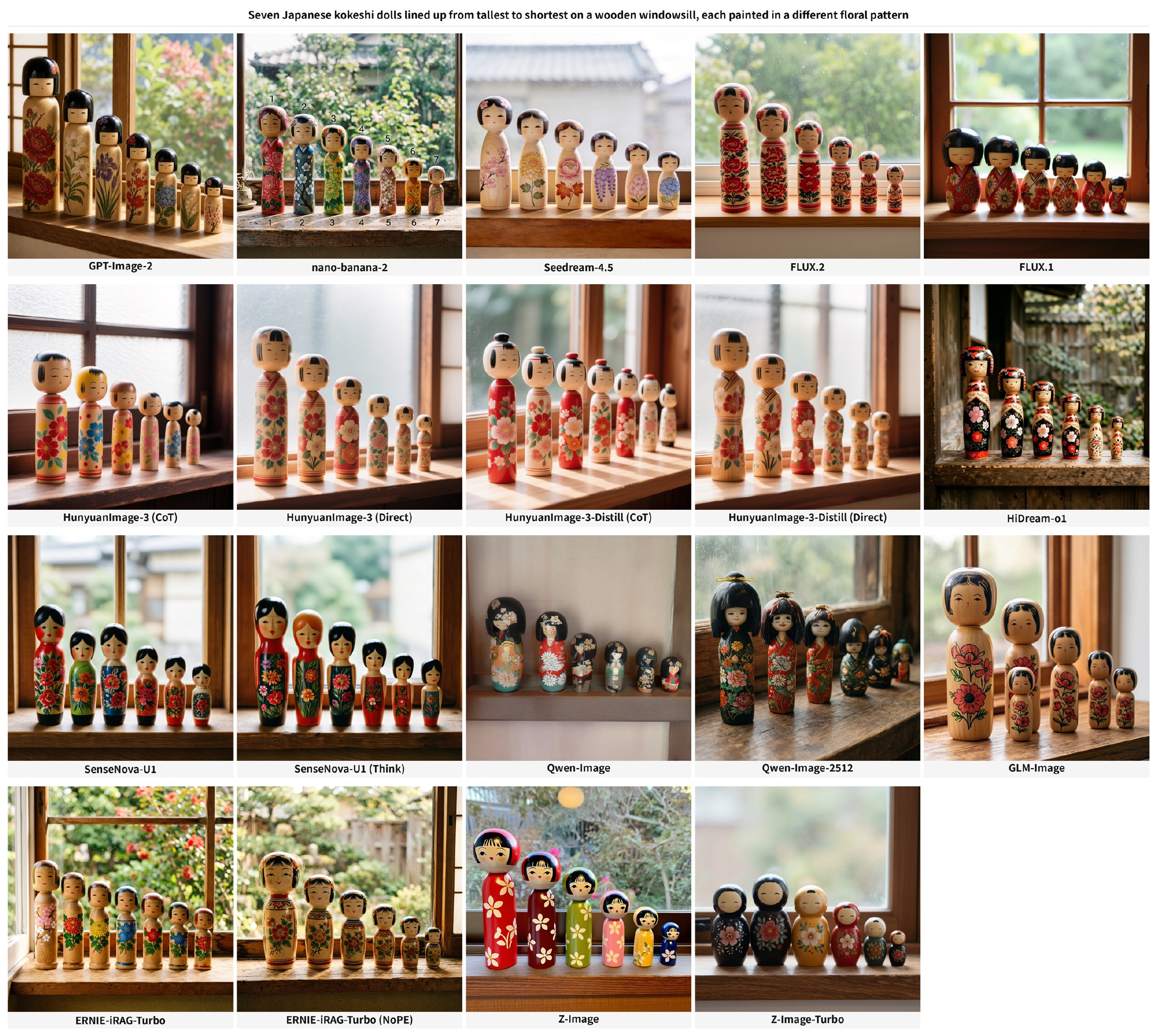}
\end{figure}

\begin{figure}[h]
  \centering
  \includegraphics[width=\linewidth]{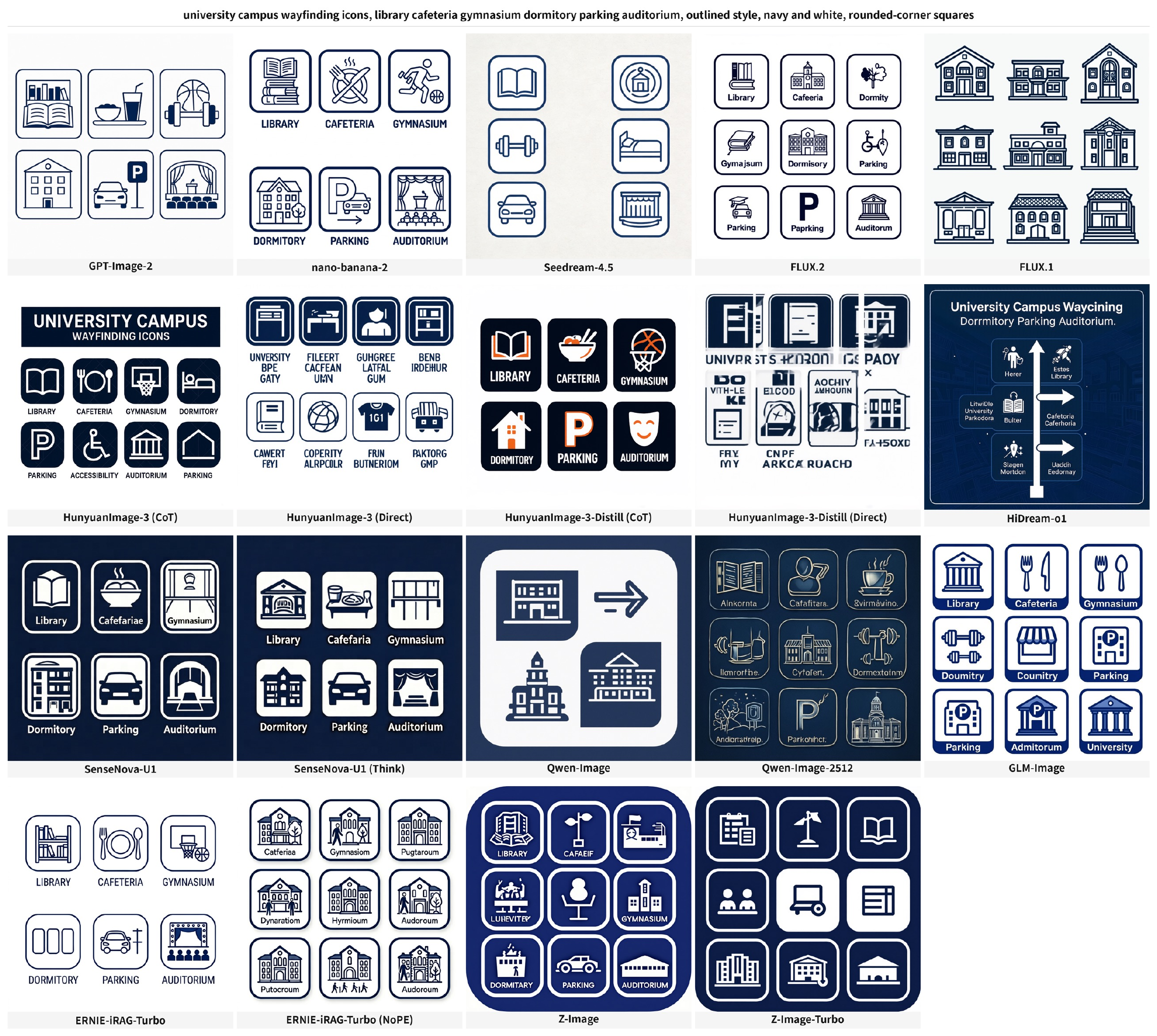}
\end{figure}

\begin{figure}[h]
  \centering
  \includegraphics[width=\linewidth]{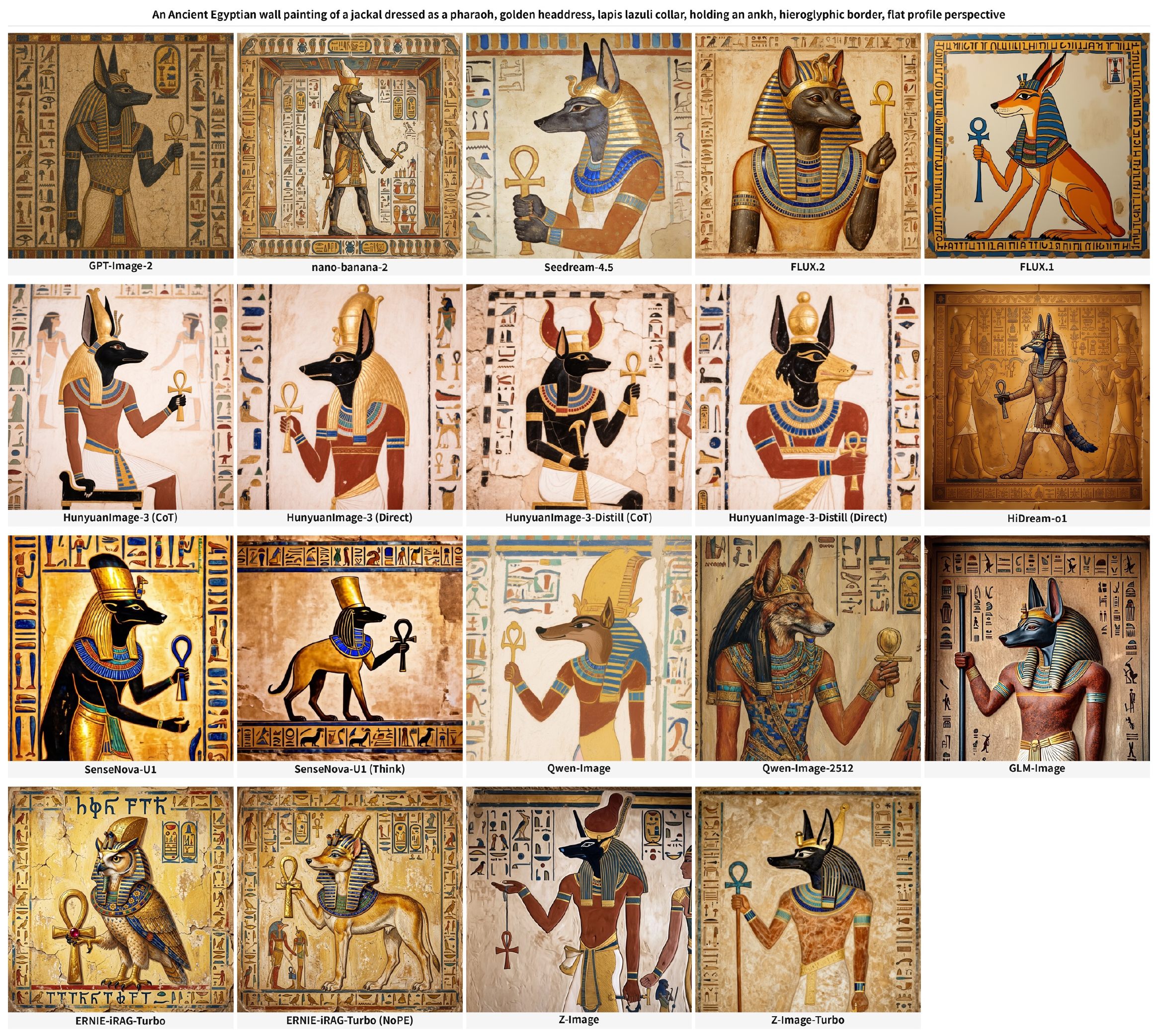}
\end{figure}

\begin{figure}[h]
  \centering
  \includegraphics[width=\linewidth]{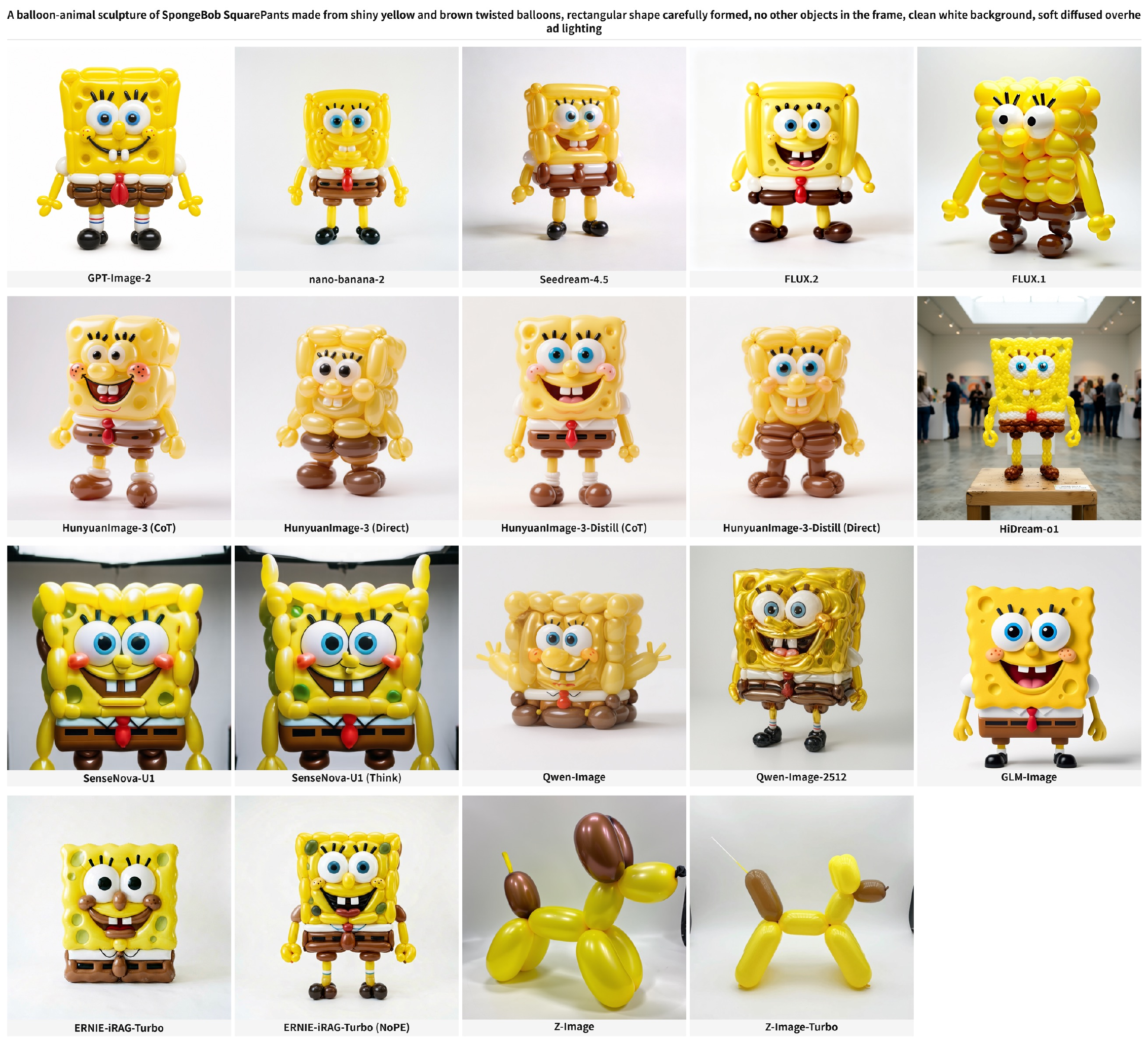}
\end{figure}

\end{document}